\def\year{2021}\relax
\documentclass[letterpaper]{article} 
\usepackage[switch]{lineno}
\usepackage{aaai21}  
\usepackage{times}  
\usepackage{helvet} 
\usepackage{courier}  
\usepackage[hyphens]{url}  
\usepackage{graphicx} 
\usepackage{natbib}  
\usepackage{caption} 
\usepackage{pifont}
\usepackage{subcaption}
\usepackage{booktabs}
\usepackage{multirow}
\usepackage{tablefootnote}  
\usepackage{footmisc}  

\newcommand{\cmark}{\ding{51}}

\newcommand{\argmax}{\mathop{\rm arg~max}\limits}
\title{Towards Fully Automated Manga Translation}
\makeatletter
    \renewcommand{\copyright@on}{F}
\makeatother
\author {
        Ryota Hinami\textsuperscript{\rm 1*},
        Shonosuke Ishiwatari\textsuperscript{\rm 1*},
        Kazuhiko Yasuda\textsuperscript{\rm 2},
        Yusuke Matsui\textsuperscript{\rm 2} \\
}
\affiliations {
    \textsuperscript{\rm 1} Mantra Inc. \\
    \textsuperscript{\rm 2} The University of Tokyo \\
}
\begin{document}

\maketitle

\begin{abstract}
We tackle the problem of machine translation (MT) of manga, Japanese comics.
Manga translation involves two important problems in MT: context-aware and multimodal translation.
Since text and images are mixed up in an unstructured fashion in manga,
obtaining context from the image is essential for its translation.
However, it is still an open problem how to extract context from images and integrate into MT models.
In addition, corpus and benchmarks to train and evaluate such models are currently unavailable.
In this paper, we make the following four contributions that
establish the foundation of manga translation research.
First, we propose a multimodal context-aware translation framework.
We are the first to incorporate context information obtained from manga images.
It enables us to translate texts in speech bubbles
that cannot be translated without using context information (e.g., texts in
other speech bubbles, gender of speakers, etc.).
Second, for training the model, we propose the approach to automatic
corpus construction from pairs of original manga and their translations,
by which a large parallel corpus can be constructed without any manual labeling.
Third, we created a new benchmark to evaluate manga translation.
Finally, on top of our proposed methods,
we devised a first comprehensive system for fully automated manga translation.

\end{abstract}
\renewcommand{\thefootnote}{\fnsymbol{footnote}}
\footnotetext[1]{Both authors contributed equally to this work.}
\insert\footins{\noindent\footnotesize Copyright \copyright 2021, Association for the Advancement of Artificial Intelligence (www.aaai.org). All rights reserved.}
\renewcommand{\thefootnote}{\arabic{footnote}}
\setcounter{footnote}{0}
\section{Introduction}
\label{sec:intro}
Comics are popular all over the world.
There are many different forms of comics around the world, such as manga in Japan,
webtoon in Korea, and manhua in China, all of which have their own unique characteristics.
However, due to the high cost of translation, most comics have not been translated
and are only available in their domestic markets.
What if all comics could be immediately translated into any language?
Such a panacea for readers could be made possible
by machine translation (MT) technology.
Recent advances in neural machine translation (NMT)~\cite{cho14,sutskever14,bahdanau15,wu16,vaswani17} have increased the number of
applications of MT in a variety of fields.
However, there are no successful examples of MT for comics.

\begin{figure}[t]
    \centering
    \includegraphics[width=\linewidth]{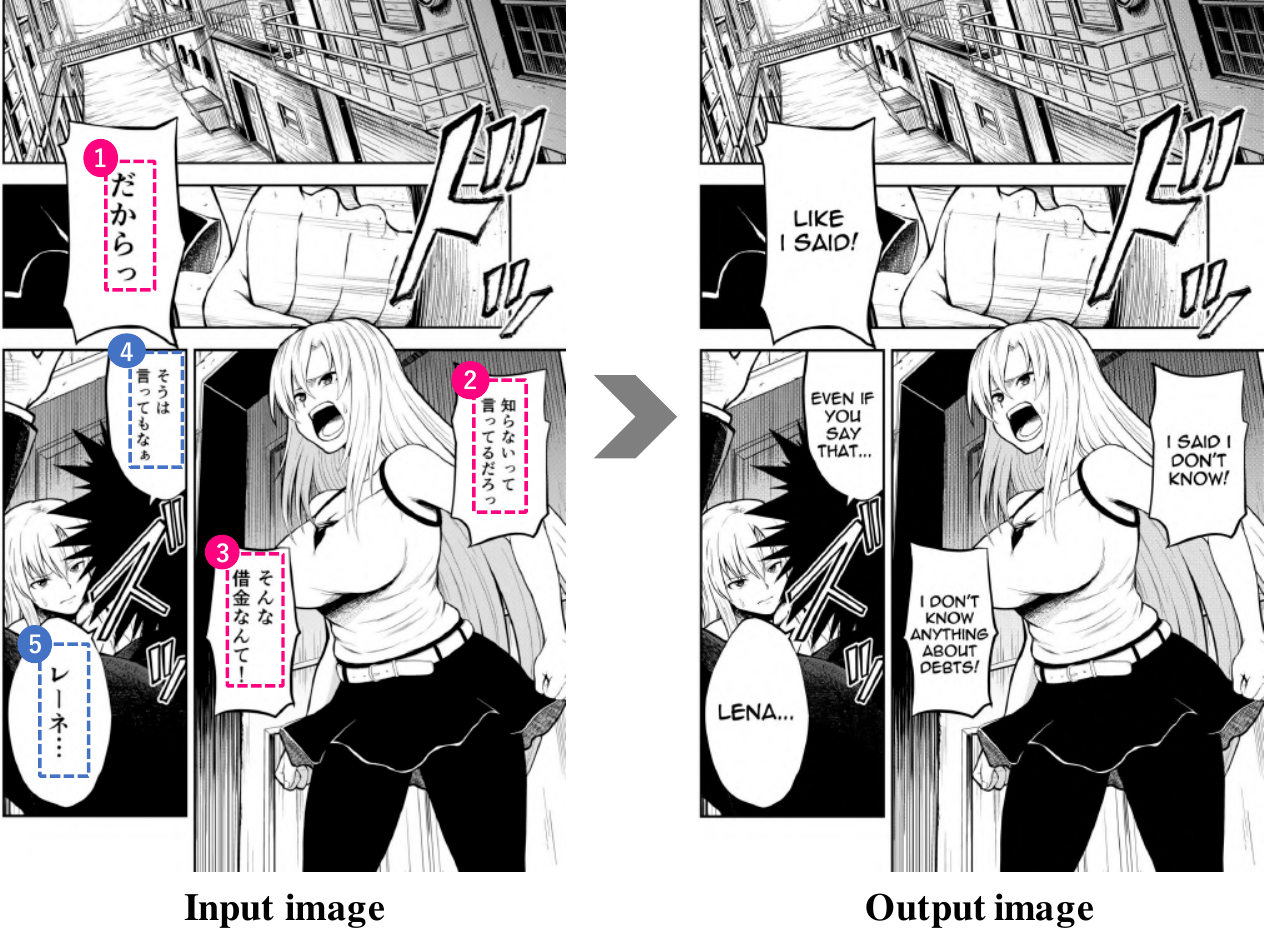}
    \caption{
        Given a manga page,
        our system automatically translates the texts on the page into English and replaces the original texts
        with the translated ones.
        \copyright Mitsuki Kuchitaka.
    }
    \label{fig:eg_first}
\end{figure}

What makes the translation of comics difficult?
In comics, an utterance by a character is often divided up into multiple bubbles.
For example, in the manga page shown on the left side of Fig.~\ref{fig:eg_first},
the female's utterance is divided into bubbles \#1 to \#3 and the male's into \#4 to \#5.
Since both the subject ``I'' and the verb ``know'' are omitted in the bubble \#3,
it is essential to exploit the context from the previous or next bubbles.
What makes the problem even more difficult is that the bubbles are not simply aligned from right to left, left to right, or top to bottom.
While the bubble \#4 is spatially closed to \#1, the two utterances are not continuous.
Thus, it is necessary to parse the structure of the manga to recognize the texts in the correct order.
In addition, the visual semantic information must be properly captured to resolve the ambiguities.
For example, some Japanese word can be translated into both ``he'', ``him'', ``she'', or ``her''
so it is crucial to capture the gender of the characters.

These problems are related to \textit{context} and \textit{multimodality};
considering context is essential in comics translation
and we need to understand an image to capture the context.
Context-aware~\cite{jean2017does,tiedemann-scherrer2017neural,wang2017exploiting} and multimodal translation~\cite{specia2016WMT,elliott2017WMT,barrault2018WMT} are both hot topics in NMT but researched independently.
Both are important in various kinds of application, such as movie subtitles or face-to-face conversations.
However, there have not been any study on how to exploit
context with multimodal information.
In addition, there are no public corpus and benchmarks for training and evaluating models,
which prevents us from starting the research of multimodal context-aware translation.

\subsection*{Contributions}
This paper addresses the problem of translating manga, meeting the grand challenge of fully automated manga translation.
We make the following four contributions that establish a
foundation for research on manga translation.

\textit{Multimodal context-aware translation.}
Our primary contribution is a context-aware manga translation framework.
This is the first approach that incorporates context
information obtained from an image into manga translation.
We demonstrated it significantly improves the performance of
manga translation and enables us to translate texts
that are hard to be translated without using context information, such as the example presented above with Fig.~\ref{fig:eg_first}.

\textit{Automatic parallel corpus construction.}
Large in-domain corpora are essential to training accurate NMT models.
Therefore, we propose a method to automatically build a manga parallel
corpus.
Since, in manga, the text and drawings are mixed up in an unstructured manner,
we integrate various computer vision techniques to extract parallel sentences from images.
A parallel corpus containing four million sentence pairs with context information
is constructed automatically without any manual annotation.

\textit{Manga translation dataset.}
We created a multilingual manga dataset, which is the first benchmark of manga translation.
Five categories of Japanese manga were collected and translated.
This dataset is publicly available.

\textit{Fully automatic manga translation system.}
On the basis of the proposed methods,
we built the first comprehensive system that translates manga fully automatically from image to image.
We achieved this capability by integrating text recognition, machine translation, and image processing into a unified system.

\section{Related Work}
\label{sec:rel}
\subsection{Context-aware machine translation}
Despite the recent rapid progress in NMT~\cite{cho14,sutskever14,bahdanau15,wu16,vaswani17}, most models are not designed to capture extra-sentential context. The sentence-level NMT models suffer from errors due to linguistic phenomena such as referential expressions (e.g., outputting `him' when correct output is `her') or omitted words in the source text~\cite{voita2019good}. There have been interests in modeling extra-sentential context in NMT to cope with these problems. The previously proposed methods aimed at context-aware NMT can be categorized into two types: (1) extending translation units from a single sentence to multiple sentences ~\cite{tiedemann-scherrer2017neural,bawden2018evaluating,scherrer2019analysing}; and (2) adding modules that capture context information to NMT models~\cite{jean2017does,wang2017exploiting,tu2018learning,werlen2018document,voita2018context,maruf2018document,zhang2018improving,maruf2019selective,xiong2019modeling,voita2019context,voita2019good}.

While the various methods have been evaluated on different language pairs and domains, we mainly focused on Japanese-to-English translation in manga domains.
Our scene-based translation is deeply related to 2+2 translation~\cite{tiedemann-scherrer2017neural},
which incorporates the preceding sentence by prepending it to be the current one.
While it captures the context in the previous sentence, our scene-based model considers all the sentences in a single scene.

\subsection{Multimodal machine translation}
The manga translation task is also related to multimodal machine translation (MMT).
The goal of the MMT is to train a visually grounded MT model by using sentences and images~\cite{harnad90,glenberg00}.
More recently, the NMT paradigm has made it possible to handle discrete symbols (e.g., text) and continuous signals (e.g., images) in a single framework~\cite{specia2016WMT,elliott2017WMT,barrault2018WMT}.

The manga translation can be considered as a new challenge in the MMT field for several reasons.
First, the conventional MMT assumes a single image and its description as inputs~\cite{elliott2016multi30k}. However, manga consists of multiple images with context, and the texts are drawn in the images. 
Second, the commonly used pre-trained image encoders~\cite{ijcv_russakovsky2015} cannot be used to encode manga images as they are all trained on natural images.
Third, no parallel corpus is available in the manga domain.
We tackled these problems by developing a novel framework to extract visual/textual information from manga images and an automatic corpus construction method.

\begin{figure*}
    \includegraphics[width=1.0\linewidth]{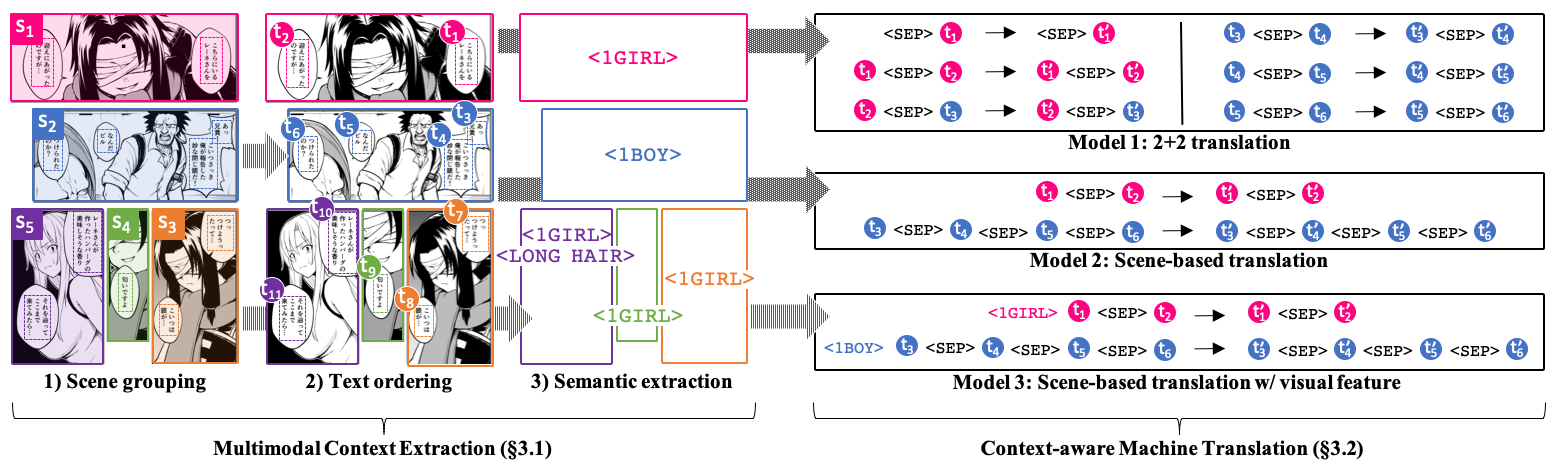}
    \caption{Proposed manga translation framework. N' represents the translation of a source sentence N. \copyright Mitsuki Kuchitaka }
    \label{fig:context_extraction}
\end{figure*}

\section{Context-Aware Manga Translation} \label{sec:model}
Now let us introduce our approach to manga translation that incorporates the multimodal context.
In this section, we will focus on the translation of {\it texts} with the help of image information, assuming that the text has already been recognized in an input image.
Specifically, suppose we are given a manga page image $I$ and $N$ unordered texts on the page.
The texts are denoted as $\mathcal{T}$, where $|\mathcal{T}|=N$.
We are also given a bounding box for each text: $\mathbf{b}(t)=[x, y, w, h]^\top$.
Our goal is to translate each text $t \in \mathcal{T}$ into another language $t'$.

The most challenging problem here is that we cannot translate each $t$ independently of each other.
As discussed in the introduction, incorporating texts in other speech bubbles is indispensable to translate each $t$.
In addition, visual semantic information such as the gender of the character sometimes helps translation.
We first introduce our approach to extracting such context from an image and then describe our translation model using those contexts.

\subsection{Extraction of Multimodal Context} \label{sec:context_extraction}
We extract three types of context, i.e., scene, reading order, and visual information,
which are all useful information for multimodal context-aware translation.
The left side of Fig.~\ref{fig:context_extraction} illustrates the three procedures explained below 1)--3).

\paragraph{1) Grouping texts into scenes:}
A single manga page includes multiple frames,
each of which represents a single scene.
In the translation of the story,
the texts included in the same scene are usually more useful for translation than the texts in a different scene.
Therefore, we group texts into scenes to determine the ones useful as contexts.
First, we detect frames in a manga page using an object detector
by regarding each frame as an object in the manner of \cite{corr_ogawa2018}.
In particular, we trained the Faster R-CNN detector~\cite{nips_ren2015} with
the Manga109 dataset~\cite{mtap_matsui2017}.
Given a manga page, the detector outputs
a set of scenes $\mathcal{S}$.
Each scene $\mathbf{s} \in \mathcal{S}$ is represented as a bounding box of
a frame: $\mathbf{s}=[x, y, w, h]^\top$.
For each text $t \in \mathcal{T}$, we find the scene $\mathbf{s} \in \mathcal{S}$
that the text belongs to.
Such a scene is defined as one that maximally overlaps the bounding box of the text.
This is determined by an assignment function $\mathbf{a}: \mathcal{T} \to \mathcal{S}$, where
$\mathbf{a}(t) = \argmax_{\mathbf{s} \in \mathcal{S}} \mathrm{IoU}(\mathbf{b}(t), \mathbf{s})$,
where $\mathrm{IoU}$ computes the intersection over the union for two boxes.

\paragraph{2) Ordering texts:}
Next, we estimate the reading order of the texts.
More formally, we sort the unordered set $\mathcal{T}$ to make an ordered set $\{t_1, \dots, t_N\}$
as shown in the left side of Fig.~\ref{fig:context_extraction}.
Since, in manga, a single sentence is usually divided up into multiple text regions,
it is quite important to ensure the text order is correct.
Manga is read on a frame-by-frame basis.
Therefore, the reading order of the texts is determined from
the order of 1) the frames and 2) the texts in each frame.
We estimate the order of the frames from the general structure of manga:
\textit{each page consists of one or more rows,
each consisting of one or more columns, recursively repeating.}
Each page is read sequentially from the top row,
and each row is read from the right column.
On the basis of this knowledge, we estimate the reading order
by recursively splitting mange page vertically and horizontally.
Afterward, the reading order of the texts in each frame is determined by the distance from the upper right point of each frame.
Even though this approach does not use any supervised information,
it accurately estimates the reading order of the frames.
Some examples are shown in Fig.~\ref{fig:text_order}
We confirmed that it could identify the correct reading order of 
91.9\% of the 258 pages we tested
(we evaluate with PubManga dataset introduced in the experiments section).
The remaining 8.2\% were irregular cases
(e.g., diagonally separated, multiple frames overlapping, etc.).

\begin{figure}[t]
    \begin{tabular}{cc}
        \begin{minipage}[t]{0.46\hsize}
            \centering
            \includegraphics[width=1.0\linewidth]{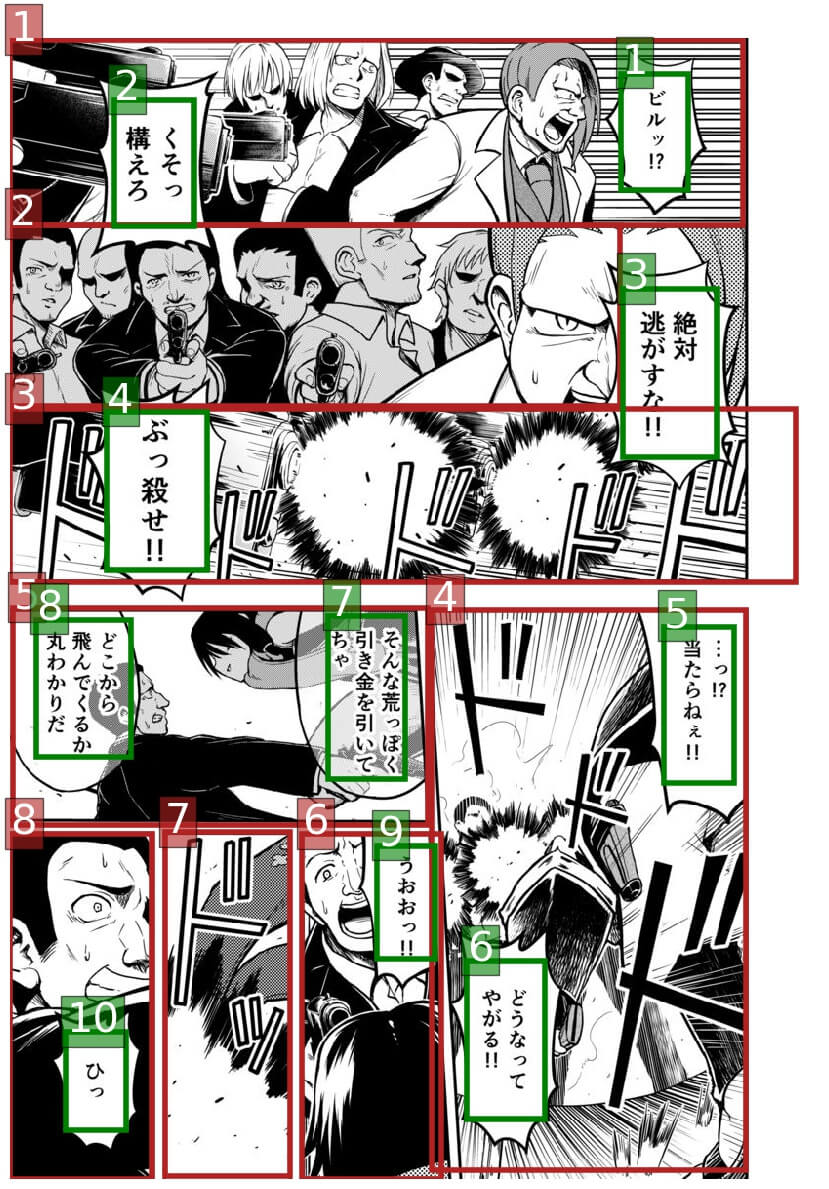}
            \label{label1}
        \end{minipage}
        \hspace{0.2cm}
        \begin{minipage}[t]{0.46\hsize}
            \centering
            \includegraphics[width=1.0\linewidth]{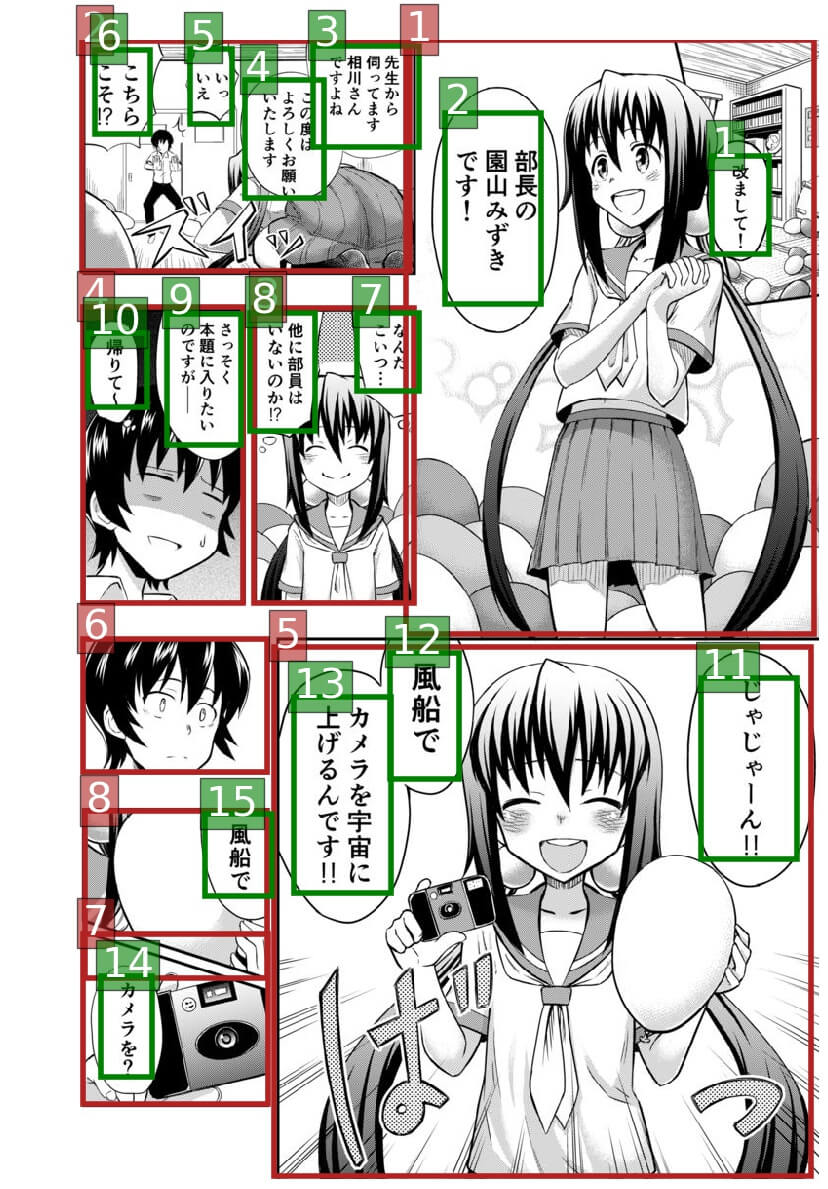}
            \label{label2}
        \end{minipage}
    \end{tabular}
    \caption{Results of our text and frame order estimation.
    The bounding boxes of text and frame are shown in the green and red rectangles, respectively.
    The estimated orders of text and frame are depicted at the upper left corner of bounding boxes.
        \copyright Mitsuki Kuchitaka}
    \label{fig:text_order}
\end{figure}

\paragraph{3) Extracting visual semantic information:}
Finally, we extract visual semantic information, such as the objects appearing in the scene.
To exploit the visual semantic information in each scene,
we predict semantic tags for each scene by using the illustration2vec model~\cite{saito2015illustration2vec}.
Given a target scene $\mathbf{s} \in \mathcal{S}$,
the illustration2vec module $f$ describes the scene by predicting semantic tags:
$f(\mathbf{s}) \subseteq \mathcal{L}$.
In the illustration2vec model,
$\mathcal{L}$ contains 512 pre-defined semantic tags:
$\mathcal{L} = \{\mathtt{1GIRL}, \mathtt{1BOY}, \dots   \}$.
Several tags can be predicted from a single scene.
Although we tried integrating a deep image encoder
as is done in many multimodal tasks~\cite{zhou2018visual,fukui2016multimodal,vinyals2015show},
it did not improve performance on our tasks.

\paragraph{}
We should emphasize that this framework is not limited to manga.
It can be extended to any kind of media having multimodal context,
including movies and animations, by properly defining the scene.
For example, it can be easily applied to movie subtitle translation
by extracting contexts in three steps: 1) segmenting videos into scenes,
2) ordering texts by time,
and 3) extracting semantic tags by video classification.

\subsection{Context-aware Translation Model} \label{sec:translation_model}
To incorporate the extracted multimodal context into MT, we take a simple yet effective concatenation approach~\cite{tiedemann-scherrer2017neural,junczys-dowmunt2019microsoft}: concatenate multiple continuous texts and translate them with a sentence-level NMT model all at once.
Note that any NMT architecture can be incorporated with this approach.
In this study, we chose the Transformer (big) model and set its default parameters in accordance with~\cite{vaswani17}.
The right side of Fig.~\ref{fig:context_extraction} illustrates the three models explained below.

\paragraph{Model1: 2+2 translation.}
The simplest method utilizes the previous text as context. To train and test the model, we prepend the previous text in the source and target languages~\cite{tiedemann-scherrer2017neural}. That is, to translate $t_n$ into $t'_n$, two texts $t_{n-1}$ and $t_n$ are fed into the translation model, which outputs $t_{n-1}'$ and $t'_n$. The boundary of the two texts is marked with a special token \texttt{<SEP>}.

\paragraph{Model2: Scene-based translation.}
Considering only the previous text as the context is not always sufficient. We may want to consider two or more previous texts or even the subsequent texts in the same scene. To enable this, we generalize the 2+2 translation by concatenating all the texts in each frame and translating them all at once. This procedure is illustrated as follows. Suppose we would like to translate $t_n$ to $t'_n$. Unlike Model1 that makes use of $t_{n-1}$ only, we feed $\{t \in \mathcal{T} \mid \mathbf{a}(t_n) = \mathbf{a}(t)\}$ into the model.

\paragraph{Model3: Scene-based translation with visual feature}
To incorporate the visual information into Model2, we prepend the predicted tags to the sequence of the input texts. Each tag is represented as a special token, such as \texttt{<1GIRL>} or \texttt{<1BOY>}. Note that this does not lead to any changes in the model itself. By adding the tags as input, we let the model consider the visual information when needed. This means that, to translate $t_n$ into $t'_n$, we additionally input $f(\mathbf{a}(t_n))$.

\section{Parallel Corpus Construction} \label{sec:corpus}

We propose the approach to automatic corpus construction for training our translation model.
Given a pair of manga books as input: a Japanese manga and its English-translation,
our goal is to extract parallel texts with context information
that can be used to train the proposed model.
This is a challenging problem because manga is regarded as a sequence of images without any text data.
Since texts are scattered all over the image and are written in various styles,
it is difficult to accurately extract texts and group them into sentences.
In addition, even when sentences are correctly extracted from manga images,
it is difficult to find the correct correspondence between sentences in different languages.
The differences in text direction from one language to another
(e.g., vertical in Japanese and horizontal in English) makes this problem harder.
We solve this problem by using computer vision techniques
by fully utilizing
the structural features of manga images, such as the pixel-level locations of the speech bubbles.

\begin{figure}
    \centering
    \includegraphics[width=0.9\linewidth]{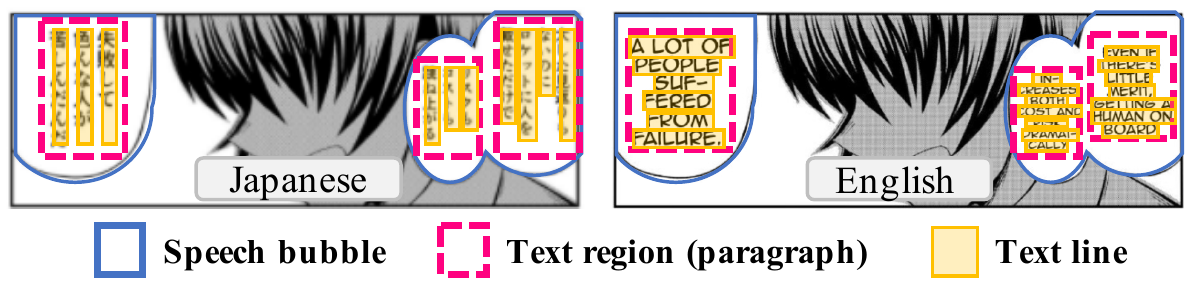}
    \caption{Term~definition for manga text. \small{\copyright Mitsuki Kuchitaka}}
    \label{fig:term_definition}
\end{figure}

\paragraph{Terms and available labeled data}
First though, let us define the terms associated with manga text;
Fig.~\ref{fig:term_definition} illustrates speech bubbles, text regions, and text lines.
One speech bubble contains one or more text regions (i.e., paragraph),
each comprising one or more text lines.
We assume that only the annotation of speech bubbles is available for training models;
annotations of text lines and text regions are unavailable.
In addition, segmentation masks of speech bubbles and any data in the target language are also unavailable.
This is a natural assumption because current public datasets only have
speech bubble-level bounding box annotations of the Japanese version for manga~\cite{mtap_matsui2017}
and those of English version for American-style comics~\cite{cvpr_iyyer2017,icdar_guerin2013}.
This limitation on labeled data is one of the challenges of
parallel text extraction from comics.
Note that our approach does not depend on specific languages.
We also applied it to Chinese as a target language in addition to English,
which is demonstrated later in Fig.~\ref{fig:e2e_example}.

\begin{figure*}
    \centering
    \includegraphics[width=\linewidth]{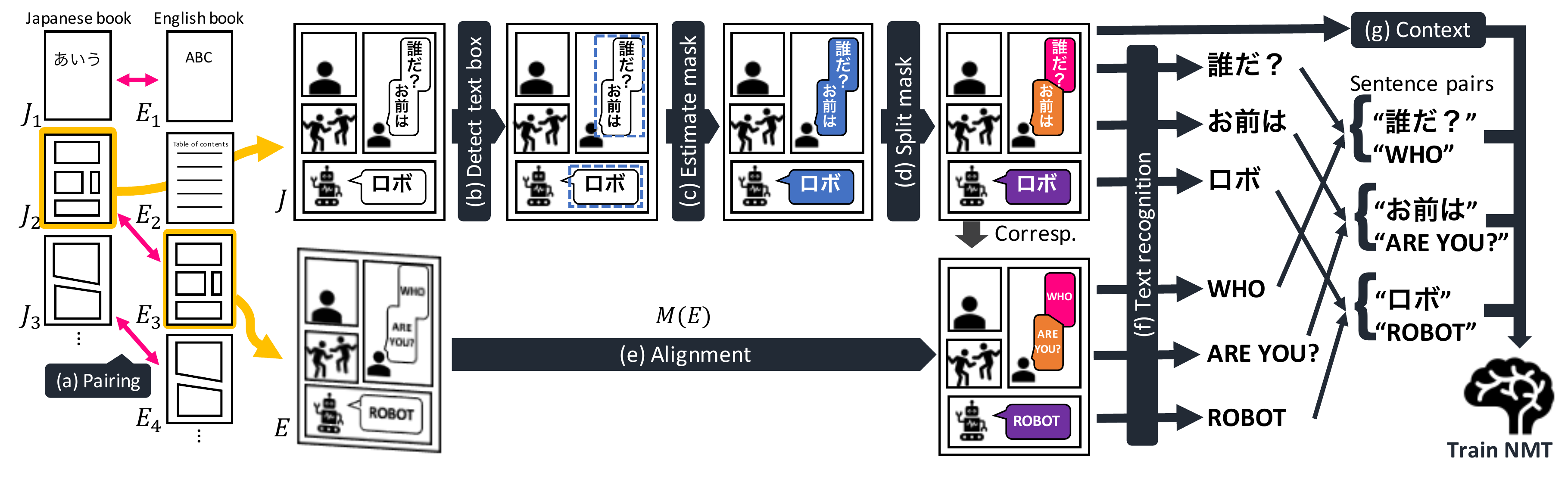}
    \caption{Proposed framework of parallel corpus construction.}
    \label{fig:framework_collecting_groundtruth}
\end{figure*}

\begin{figure}
    \includegraphics[width=0.95\linewidth]{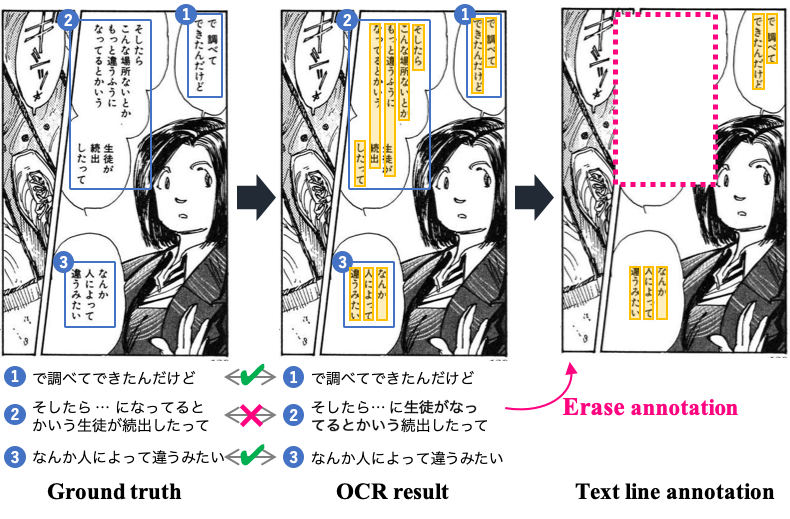}
    \caption{Generation of textline annotations.\copyright Yasuyuki Ohno}
    \label{fig:textline_detector}
\end{figure}

\paragraph{Training of Detectors} \label{sec:train_det}
We train two object detectors: speech bubble and text line detectors,
which is the basic building block of our corpus construction pipeline.
We use Faster R-CNN model with ResNet101 backbone~\cite{cvpr_he2016} for both object detectors.
The object detectors are trained with the annotation of bounding boxes.
While the speech bubble detector could be trained with public datasets (e.g., Manga 109),
the annotations of the text lines were not available.
Therefore, we devised a way to generate annotations of text lines
from the speech bubble-level annotation in a weakly supervised manner.
Fig.~\ref{fig:textline_detector} illustrates the process of 
generating annotations.
Suppose we have images with annotations of the speech bubbles'
bounding boxes and texts. In this paper, we use the annotations of Manga109
dataset~\cite{corr_ogawa2018,aizawa2020building}.
We detect the text line with a rules-based approach
(whose algorithm is described in supplementary material);
then we recognize characters using our text line recognition module.
If the recognized text perfectly matches the ground truth,
we consider the text lines to be correct and use them as the annotation of text lines.
Other text regions that were not recognized are filled in with white
(e.g., inside the dotted rectangle in Fig.~\ref{fig:textline_detector}).
The object detector is trained with the generated images and bounding box annotations.
Although the rules-based approach sometimes misses complicated patterns
such as bubbles containing multiple text regions,
the object detector can detect them by capturing 
the intrinsic properties of text lines.

\subsection{Extraction of Parallel Text Regions} \label{sec:extract_text}

Fig.~\ref{fig:framework_collecting_groundtruth} (a)--(g) illustrates the proposed pipeline
for extracting parallel text regions.

\textit{(a) Pairing pages.}
Let us define an input Japanese manga as a set of $n_j$ images (pages), denoted as $\{J_1, \dots, J_{n_j}\}$.
Similarly, let us define the English manga as a set of $n_e$ pages: $\{E_1, \dots, E_{n_e}\}$.
Note that typically $n_j \ne n_e$,
because pages such as the front cover, table of contents, and illustration
can be optionally included or removed during the production of the translation.
Owing to such inconsistencies, we must find \textit{page-wise correspondences} first
as shown in Fig.~\ref{fig:framework_collecting_groundtruth} (a).
We find the correspondences by global descriptor-based image retrieval combined with spatial verification~\cite{radenovic2018revisiting}.
For each Japanese page $J_i$, we first retrieve the English image $E_j$ with the highest similarity to $J_i$ from $\{E_1, \dots, E_{n_e}\}$,
where the similarity of two pages is computed as the $L_2$ distance of global features extracted by the deep image retrieval (DIR) model~\cite{gordo2016deep}. 
We then apply spatial verification~\cite{philbin2007object} to reject false matching pairs. 
The homography matrix between two pages is estimated by RANSAC~\cite{fischler1981random} with AKAZE descriptors~\cite{bmvc_alcantarilla2013}.
If the number of inliers in RANSAC is more than 50, we decide that $J_i$ and $E_j$ are corresponding.

\textit{(b) Detection of text boxes.}
After the page-aligning step, we obtain a set of corresponding pairs of English and Japanese pages. Hereafter, we discuss how to extract a parallel corpus from a single pair, $J$ and $E$.
First, the bounding boxes of the speech bubbles are obtained by applying the speech bubble detector to $J$.

\textit{(c) Pixel-level estimation of speech bubbles.}
We estimate the precise pixel-level mask for each bubble from the bounding box.
We employ edge detection with canny detector~\cite{canny1986computational}
to detect the contour of speech bubbles.
For each bounding box of a speech bubble,
we select the connected component of non-edge pixels that shares the largest area with the bounding box,
which is the blank area inside the speech bubble.
In this way, we precisely estimate the masks of the speech bubbles without having to worry about how to train a semantic segmentation model that cannot be trained with the currently available dataset.

\textit{(d) Splitting connected speech bubbles.}
As illustrated in Fig.~\ref{fig:term_definition},
sometimes a speech bubble includes multiple text regions (i.e., paragraphs).
We split up such speech bubbles in order to identify the text regions by clustering the text lines.
The text lines obtained by the object detector are then grouped into paragraphs by clustering
the vertical coordinates at the top of text lines with MeanShift~\cite{comaniciu2002mean}.
Finally, masks are split so that all text regions are perfectly separated, and the length of the boundary (i.e., splitting length) is minimized.

\textit{(e) Alignment between languages.}
We then estimate the masks of text regions for $E$ by aligning $J$ and $E$.
Since the scales and margins are often different between $J$ and $E$,
$E$ is transformed so that the two images overlap exactly.
We update $E$ by applying a perspective transformation: $E \gets M(E)$,
where $M(\cdot)$ indicates
the transformation computed in the previous page pairing step.
The resulting page has a better pixel-level alignment so that text regions
in $E$ can be easily localized from the text regions in $J$.
Such a correspondence is made possible by the distinctive nature of manga:
the translated text is located in the same bubble.
Note that we do not use any learning-based models for $E$ in  steps 1)--5),
so our method can be used for any target language
even if a dataset for learning detectors is unavailable.

\textit{(f) Text recognition.}
Given the segmentation masks of the text regions,
we recognize the characters for each image pair $J$ and $E$.
Since we found that existing OCR systems perform very poorly on manga text
due to the variety of fonts and styles that are unique to manga text,
we developed an OCR module optimized for manga.
We developed our own text rendering engine that generates text images optimized for manga.
Five millions of text images are generated with the engine,
by which we train the OCR module based on the model of Baek et al.~\cite{baek19}.
Technical details of this component are described in the supplementary material.

\textit{(g) Context extraction.}
We extract the context information
(i.e., the reading order and scene labels of each text)
from $J$ in the manner described in the previous section.

\section{Experiments}
\label{sec:exp}
\subsection{Dataset}
Although there are no manga/comics datasets comprising of multiple languages,
we created two new manga datasets, i.e., \textbf{OpenMantra} and \textbf{PubManga},
one to evaluate the MT, the other to evaluate the constructed corpus.

\paragraph{\textbf{OpenMantra}:}
While we need a ground-truth dataset to evaluate the NMT models, no parallel corpus in the manga domain is available.
Thus, we started by building OpenMantra, an evaluation dataset for manga translation.
We selected five Japanese manga series across different genres, including fantasy, romance, battle, mystery, and slice of life.
In total, the dataset consists of 1593 sentences, 848 frames, and 214 pages. After that, we asked professional translators to translate the whole series into English and Chinese.
This dataset is publicly available for research purposes.\footnote{https://github.com/mantra-inc/open-mantra-dataset}

\paragraph{\textbf{PubManga}:}
OpenMantra is not appropriate for evaluating the constructed corpus because translated versions are created by ourselves.
Thus, we selected nine Japanese manga series across different categories, each having 18--40 pages (258 pages in total), and created another dataset of published translations (PubManga).
This dataset includes annotations of 1) bounding boxes of the text and frame, 2) texts (character sequence) in both Japanese and English, and 3) the reading order of the frames and texts.
The annotations and full list of manga titles are available upon request.

\begin{table*}[t]
    \centering
        \caption{System description and translation performances on the OpenMantra Ja--En dataset. * indicates the result is significantly better than Sentence-NMT (Manga) at $p < 0.05$.}
        \label{tab:result_bleu}
        \begin{tabular}{@{}l|ll|rr@{}}
            \toprule
            System & Training corpus & Translation unit & Human & BLEU \\
            \midrule
            Without context & & \\
            ~~Google Translate\tablefootnote{We ran the test on Apr. 17, 2020.} & N/A & sentence & - & 8.72 \\
            ~~Sentence-NMT (OS18) & OpenSubtitles2018 & sentence & 2.11 & 9.34 \\
            ~~Sentence-NMT (Manga) & Manga Corpus & sentence & 2.76 & 14.11 \\
            \midrule
            With context & & \\
            ~~2 + 2 ~\cite{tiedemann-scherrer2017neural} & Manga Corpus & 2 sentences & 2.85 & 12.73 \\
            ~~Scene-NMT & Manga Corpus & frame& \textbf{2.98}* & 12.65 \\
            ~~Scene-NMT w/ visual & Manga Corpus & frame & \textbf{2.91}* & 12.22\\
            \bottomrule
        \end{tabular}
\end{table*}

\subsection{Evaluation of Machine Translations} \label{sec:eval_translation}

To confirm the effectiveness of our models and Manga corpus, we ran translation experiments on the OpenMantra dataset.

\paragraph{Training corpus:}
To train the NMT model for manga, we collected training data by the proposed corpus construction approach.
We prepared 842,097 pairs of manga pages that were published in both Japanese and English.
Note that all the pages are in digital format without textual information.
3,979,205 pairs of Japanese--English sentences were obtained automatically.
We randomly excluded 2,000 pairs for validation purposes.

In addition, we used OpenSubtitles2018 (OS18)~\cite{lison18}, a large-scale parallel corpus to train a baseline model.
Most of the data in OS18 are conversational sentences extracted from movie and TV subtitles, so they are relatively similar to the text in manga.
We excluded 3K sentences for the validation and 5K for the test and used the remaining 2M sentences for training.

\paragraph{Methods:}
Table~\ref{tab:result_bleu} shows the six systems used in our evaluation.
\textbf{Google Translate}\footnote{\url{https://translate.google.com/}} is an NMT system used in several domains, but the sizes and domains of its training corpus have not been disclosed.
We chose the \textbf{Sentence-NMT (OS18)} as another baseline.
The model is trained with the OS18 corpus; therefore, there are no manga domain texts included in its training data.
The \textbf{Sentence-NMT (Manga)} was trained on our automatically constructed Manga corpus described in the previous section.
\textbf{Sentence-NMT (OS18)} and \textbf{Sentence-NMT (Manga)} use the same sentence-level NMT model.

While the first three systems are sentence-level NMTs, the fourth to sixth ones are proposed context-aware NMT models.
We set \textbf{2 + 2}~\cite{tiedemann-scherrer2017neural} (Model1) as the baseline and compared their performance with those of our \textbf{Scene-NMT} models with and without visual features (Model3 \& Model2, respectively).

\paragraph{Evaluation procedure:}
Manga translation differs from plain text translation because the content of the images influences the ``feeling'' of the text.
To examine how readers actually feel when reading a translated page, we conducted a manual evaluation of translated pages instead of plain texts.
We recruited En--Ja bilingual manga readers.
They were given a Japanese page and translated English ones, and they were asked to evaluate the quality of the translation of each English page.
Following the procedure in the Workshop on Asian Translation~\cite{nakazawa18}, we asked five participants to score the texts from 1 (worst; less than 20\% of the important information is correctly translated) to 5 (best; all important information is correctly translated).
All the methods explained above other than \textbf{Google Translate} were compared. The order of presenting the methods was randomized.
In total, we collected 5 participants $\times$ 5 methods $\times$ 214 pages $=5,350$ samples.
See the supplemental material for the details of the evaluation system.
We also conducted an automatic evaluation using the BLEU~\cite{papineni02}.

\paragraph{Results:}

Table~\ref{tab:result_bleu} shows the results of the manual and automatic evaluation. The huge improvement of the \textbf{Sentence-NMT (Manga)} over \textbf{Google Translate} and \textbf{Sentence-NMT (OS18)} indicates the effectiveness of our strategy of Manga corpus construction.

A pair-wise bootstrap resampling test~\cite{koehn04} on the results of the human evaluation shows that the \textbf{Scene-NMT} outperformed the \textbf{Sentence-NMT (Manga)}.
On the other hand, there is no statistically significant difference between \textbf{2 + 2} and \textbf{Sentence-NMT (Manga)}.
These results suggest that not only the contextual information but also the appropriate way to group them is essential for accurate translation.

\begin{figure}
    \includegraphics[width=1.0\linewidth]{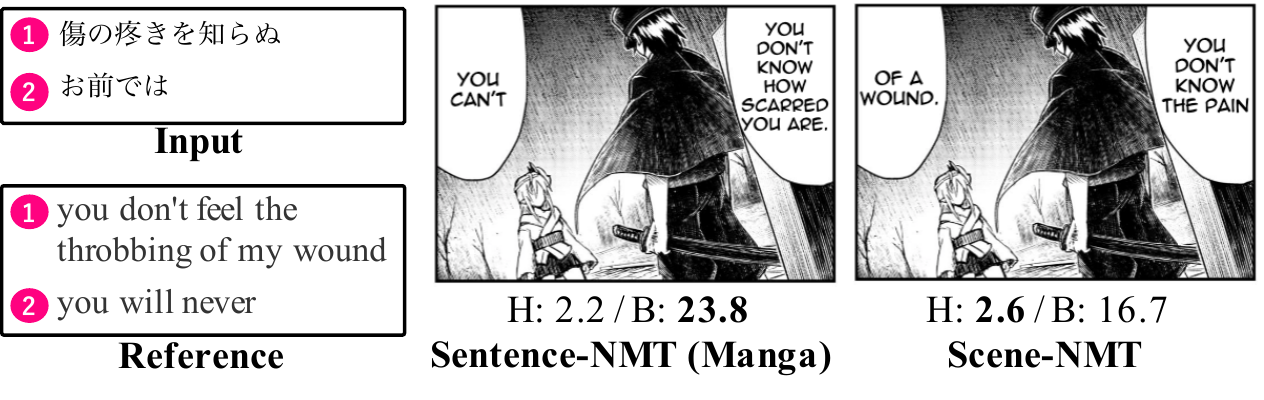}
    \caption{Outputs of the sentence-based (center) and frame-based (right) models. The values after \textbf{H} and \textbf{B} are respectively the human evaluation and BLEU scores for each page. \copyright Mitsuki Kuchitaka}
    \label{fig:eg_swapped}
\end{figure}

In contrast to the results of the human evaluation, the BLEU scores of the context-aware models (fourth to sixth lines in Table~\ref{tab:result_bleu}) are worse than that of \textbf{Sentence-NMT (Manga)}.
These results suggest that the BLEU is not suitable for evaluating manga translations.
Fig.~\ref{fig:eg_swapped} shows an example where the \textbf{Scene-NMT} outperformed \textbf{Sentence-NMT (Manga)} in the manual evaluation but had lower BLEU scores.
Here, we can see that only the \textbf{Scene-NMT} has swapped the order of the texts.
This flexibility naturally resolves the differences in word order between Japanese and English.
However, it results in a worse BLEU score since the references usually maintain the original order of the texts.

\begin{figure}
    \includegraphics[width=1.0\linewidth]{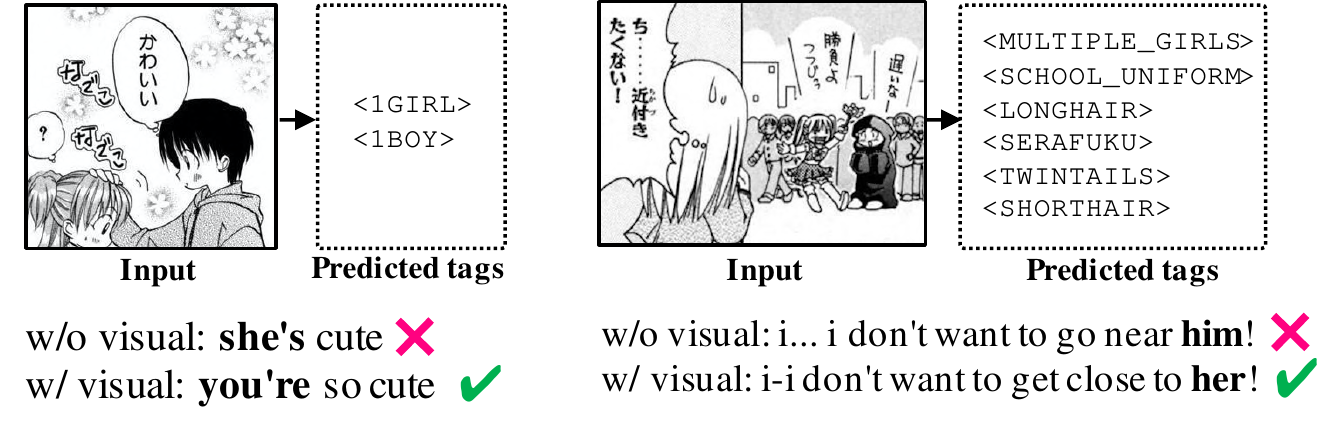}
    \caption{Translation results with and without visual features.
        \copyright Miki Ueda, \copyright Satoshi Arai.
    }
    \label{fig:eg_visual_2}
\end{figure}
\begin{figure*}
    \includegraphics[width=1.0\linewidth]{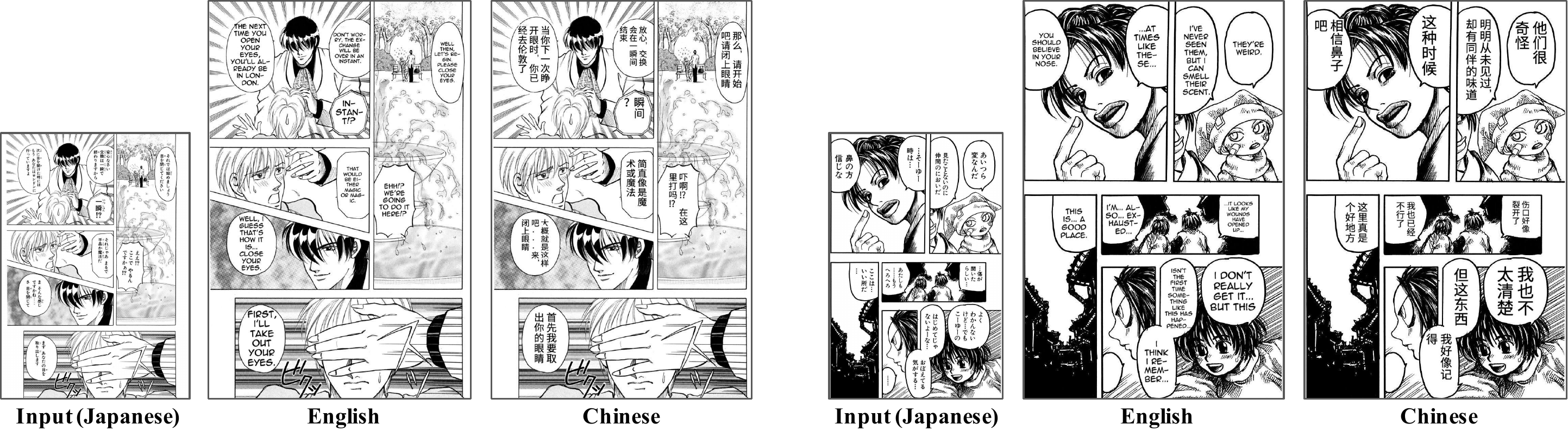}
    \caption{Results of fully automatic manga translation from Japanese to English and Chinese. \copyright Masami Taira, \copyright Syuji Takeya
    }
    \label{fig:e2e_example}
\end{figure*}

\begin{figure}
    \includegraphics[width=1.0\linewidth]{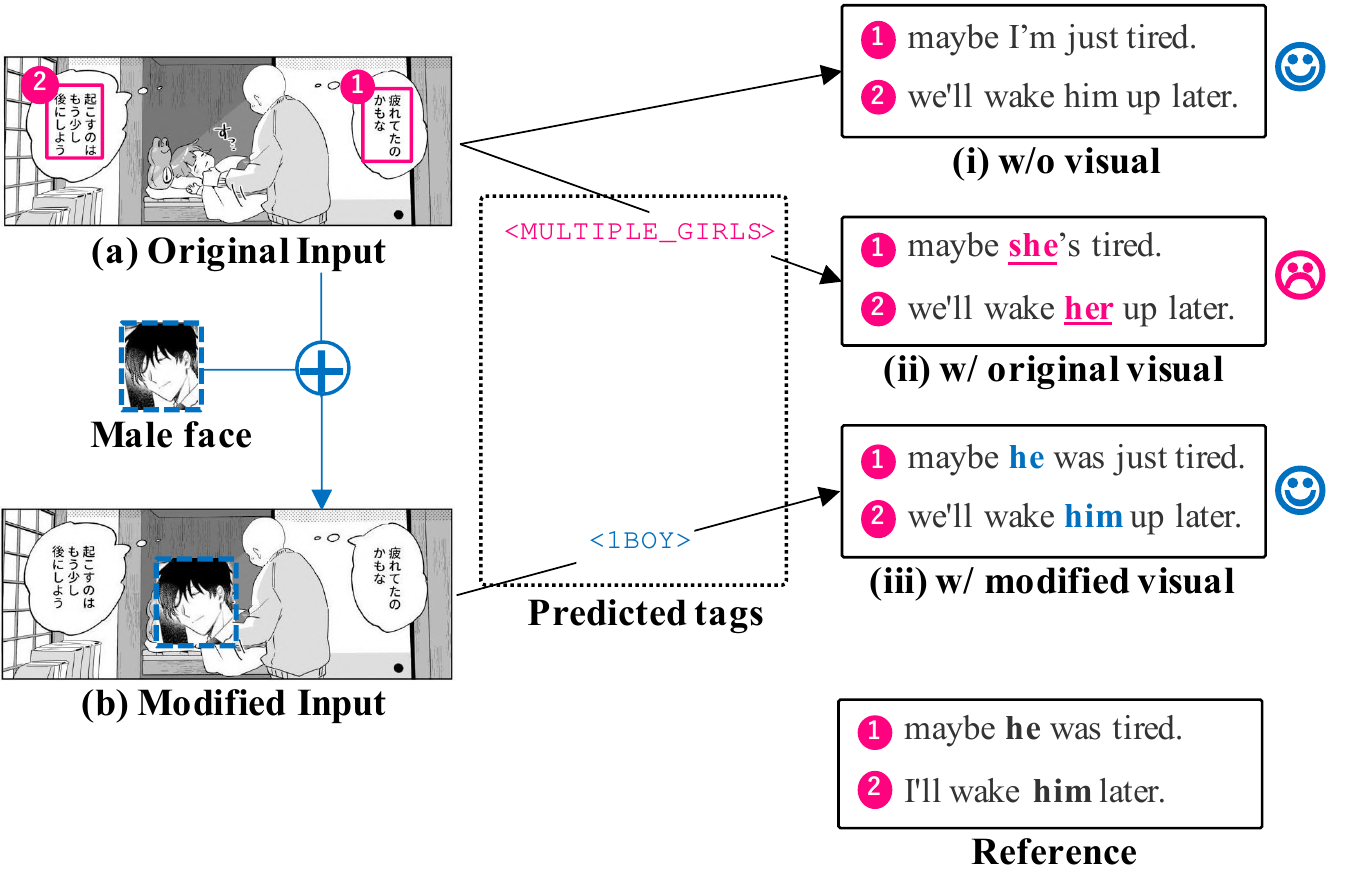}
    \caption{Translations output by the sentence-based model without (i) and with visual information (ii). By overwriting the character face in the input image (a) with a male face (b), the pronouns in the translation results (iii) are also changed. \copyright Nako Nameko}
    \label{fig:eg_visual}
\end{figure}

Although there is no statistically significant difference between \textbf{Scene-NMT} and \textbf{Scene-NMT w/ visual}, Fig.~\ref{fig:eg_visual_2} shows some promising results; pronouns (``you'' and ``her'') that cannot be estimated from textual information are correctly translated by using visual information.
These examples indicate that we need to combine textual and visual information to appropriately translate the content of manga.
However, we found that a large portion of the errors of \textbf{Scene-NMT w/ visual} are caused by the incorrect visual features.
To fully understand the impact of the visual feature
(i.e., semantic tags) on translation,
we conducted an analysis in Fig.~\ref{fig:eg_visual}:
(i) and (ii) in the figure show the outputs of the \textbf{Scene-NMT}
and \textbf{Scene-NMT w/ visual}, respectively.
The pronoun errors in (ii) are caused by the incorrect visual feature
``Multiple Girls'' extracted from the original image.
When we overwrote the character face with a male face,
\textbf{Scene-NMT w/ visual}  output the correct pronouns, as shown in (iii).
This result proved that \textbf{Scene-NMT w/ visual} model consider visual information
to determine translation results, and it would be improved
if we devise a way to extract visual features more accurately.
Designing such a good recognition model for manga images remains as future work.

\subsection{Evaluation of Corpus Construction} \label{sec:eval_corpus}
To evaluate the performance of corpus construction, we compared the following four approaches:
1) \textbf{Box}: Bounding boxes by the speech bubble detector
are used as text regions instead of segmentation masks.
This is the baseline of a simple combination of speech bubble detection and OCR.
2) \textbf{Box-parallel}: Bounding box of speech bubbles are
detected in both Japanese and English images by applying
detector to both images.
For each detected Japanese box, the English box that overlaps it most is selected as the corresponding box.
3) \textbf{Mask w/o split}: Segmentation masks of speech bubbles are estimated,
but the process of splitting the masks is not done.
4) \textbf{Mask w/ split} (the full proposed method): Segmentation masks of speech bubbles are estimated, and connected bubbles are split.
This fully utilizes the structural feature of the manga images.
In 1) and 2) the regions of bounding boxes are regarded as the
mask of text regions.

The corpus construction performances were evaluated on the PubManga dataset;
the results are listed in Tab.~\ref{tab:corpus}.
For a $>90$\% and $>70$\% match, the text pair with a normalized edit distance~\cite{karatzas2013icdar}
between the ground truth and extracted texts of more than 0.9 and 0.7
were considered true positives, respectively; this allowed for some OCR mistakes because the accuracy of the OCR module is not the main focus of this experiment.
This result shows that our approach that uses mask estimation is significantly better than
the two approaches that use only bounding-box regions.
Mask splitting also significantly improved both precision and recall.
The bounding box-based approaches fail to identify the regions of English text, especially when
the shapes of the text regions are different from those of the Japanese text;
this problem is caused by the difference in text direction.
These results indicate that parallel corpus extraction from manga cannot be
done with the simple combination of OCR and object detection;
exploiting structural information manga is effective.
Note that we use the same OCR and detection modules in these experiments.
The details of the evaluations are provided in the supplementary material.

\begin{table}
\centering
\caption{Corpus construction performance on the PubManga.}
\label{tab:corpus}
\begin{tabular}{@{}l l l l l l@{}}
    \toprule
    & \multicolumn{2}{c}{{\bf $>90$\% match }} & \multicolumn{2}{c}{{\bf $>70$\% match}} \\
    \cmidrule(lr){2-3}\cmidrule(lr){4-5}
    Method        & Recall & Prec. & Recall & Prec. \\\midrule
    Box           & 0.267 & 0.365  & 0.434 & 0.594 \\
    Box-parallel  & 0.246 & 0.614  & 0.289 & 0.722 \\
    Mask w/o split& 0.381 & 0.522  & 0.480 & 0.657 \\
    \textbf{Mask w/ split} & \textbf{0.584} & \textbf{0.653} & \textbf{0.688} & \textbf{0.769} &  \\
    \bottomrule
\end{tabular}
\end{table}

\section{Fully Automated Manga Translation System}
\label{sec:system}
We launch a fully automated manga translation system on top of
the proposed model trained with the constructed corpus.
Given a Japanese manga page, the system automatically recognizes texts,
translates them into target language,
and replaces the original texts with the corresponding translated texts.
It performs the following steps.

\textit{1) Text detection and recognition:}
Given a Japanese input page, the system recognizes texts in the same way as in the corpus construction.
This step predicts masks of the text regions and Japanese texts with their contexts.

\textit{2) Translation:}
Japanese texts are translated into the target languages by using the trained NMT model.
Since our approach to translation and corpus construction does not depend on a specific language,
we can translate the Japanese text into any target language
if unlabeled manga book pairs for constructing corpus are available.

\textit{3) Cleaning:}
The original Japanese texts are removed from the translation.
We employ an image inpainting model for this;
the regions of text lines are replaced by the inpainting model,
by which texts are removed clearly even when they are on image texture or drawing.
We used edge-connect~\cite{nazeri2019edgeconnect}, because its edge-first approach
is very good at complementing defects of drawings.

\textit{4) Lettering:}
Finally, the translated texts are rendered with optimized font size and location
on the cleaned image.
The location is one that maximizes the font size
under the condition that all texts are inside the text region.

\textit{Examples.}
Fig.~\ref{fig:e2e_example} shows the translations produced by our system.
It demonstrates that our system can automatically translate Japanese manga into English and Chinese.

\section{Conclusion \& Future Work}
We established a foundation for the research into manga translation
by 1) proposing multimodal context-aware translation method and
2) automatic parallel corpus construction, 3) building benchmarks,
and 4) developing a fully automated translation system.
Future work will look into 1) an image encoding method that can
extract continuous visual information that helps translation,
2) an extension of scene-based NMT to capture longer contexts in other scenes and pages, and
3) a framework to train the image recognition models and the NMT model jointly for more accurate end-to-end performance.

\section{Acknowledgement}
This work was partially supported by IPA Mitou Advanced Project, FoundX Founders Program, and the UTokyo IPC 1st Round Program.

The authors would like to appreciate Ito Kira, Mitsuki Kuchitaka, and Nako Nameko for providing their manga for research use, and Morisawa Inc. for providing their font data.
We also thank Naoki Yoshinaga and his research group for the fruitful discussions before the submission.
Finally, we thank the anonymous reviewers for their careful reading of our paper and insightful comments.

\bibliography{mybibtex}

\clearpage
\renewcommand{\thetable}{\Alph{table}}
\renewcommand{\thefigure}{\Alph{figure}}
\setcounter{figure}{0}
\setcounter{table}{0}
\begin{table*}[t]
\caption{Datasets used in our experiments. Annotation of Manga corpus is automatically generated by our corpus construction method.}
\label{tab:dataset}
\centering
\begin{tabular}{@{}l c c c c c c c c c@{}}
    \toprule 
& & \multicolumn{3}{c}{\bf{amount}}  & & \multicolumn{3}{c}{\bf{text annotation}} & \bf{frame}\\
\cmidrule(l){3-5}\cmidrule(l){7-9}\cmidrule(l){10-10}
Dataset     & public & \#title & \#page & \#text & \bf{translation} & characters & box & order & box\\
\midrule
Manga109~\cite{mtap_matsui2017}    & \cmark & 109 & 21,142 &  147,918&   & \cmark & \cmark &   \\
Manga corpus&        & 563 & 842,097 & 3,979,205 & En$^*$ & \cmark$^*$ & \cmark$^*$ &\cmark$^*$ & \cmark$^*$ \\
OpenMantra & \cmark & 5 & 214 & 1,593 & En, Zh & \cmark & \cmark  & & \cmark  \\
PubManga    &        & 9 & 258 & 3,152 & En & \cmark & \cmark & \cmark & \cmark \\
\bottomrule
\end{tabular}
\end{table*}

\section{Dataset details}

Table~\ref{tab:dataset} describes the details of datasets used in our experiments. 
Manga corpus is used to train our machine translation models.
OpenMantra is used to evaluate machine translation.
PubManga is used to evaluate corpus extraction and text ordering.
Manga109 is used to train and evaluate object detectors.

\section{Rule-based Text Line Detection}
\begin{figure}
    \includegraphics[width=1.0\linewidth]{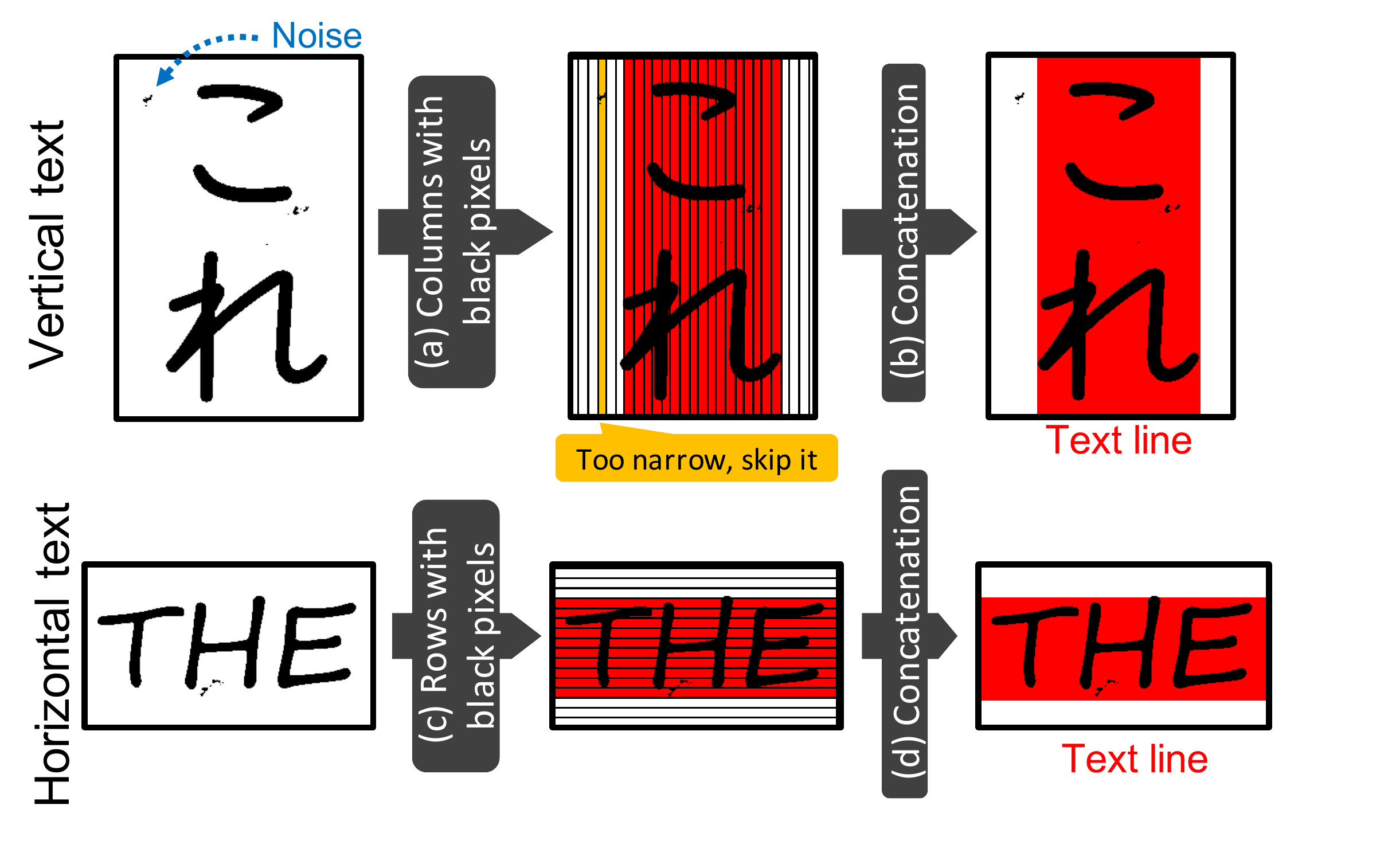}
    \caption{Rules-based text-line detection.}
    \label{fig:text_detection_recognition}
\end{figure}

Here, we introduce the rule-based approach to detect text lines without learning.
The benefit of this approach is that it does not require any labeled data.
We use this method in two parts:
1) detecting text lines of target languages (e.g., English and Chinese) in the corpus construction process
and 2) generating training data for the Japanese text line detector.
The steps of rule-based text line detection are visualized
in Fig.~\ref{fig:text_detection_recognition}~(a) and (b) for vertical texts,
and in (c) and (d) for horizontal texts.
Given a text region,
we first apply an edge detector in order to obtain connected components
that are candidates for characters (or parts of characters).
For each pixel line (a column for vertical texts, or a row in horizontal texts)
inside the text region,
we check whether connected components are included (Figs.~\ref{fig:text_detection_recognition}~(a) or (c)).
Activated consecutive columns/rows are flagged as text line candidates,
which are visualized as red or orange lines (Figs.~\ref{fig:text_detection_recognition}~(b) and (d)).
Candidates whose widths are less than half of the max widths of other candidates are removed, which removes ruby for Japanese text.

\section{Character Recognition for Manga} \label{sec:recognize_text}
Let us explain the details of text recognition of each text region,
which is described as the step 6) of Fig.~\ref{fig:framework_collecting_groundtruth}.
We found that state-of-the-art OCR systems, such as the google cloud vision API, perform very poorly on manga text due to the variety of fonts and styles that are unique to manga text.
Therefore, we developed a text recognition module optimized for manga.
The characters in each text region are recognized in two steps:
1) the text lines are detected;
2) characters on each text line are recognized.
Text lines for the Japanese image can be detected by a text line detector.
For the image of the target language, we apply a rule-based text line detection explained in the above section;
since the text regions are separated into paragraphs by a mask splitting process,
we can accurately detect text lines even with a simple rule-based approach.
Detected text lines are then fed into recognition models.
For the recognition of each text line,
we use the models trained on the data generated by our developed manga text rendering engine described below.

\begin{table*}
    \centering
    \caption{Text recognition performance on the Manga109. }
    \label{tab:text_recog}
    \begin{tabular}{@{}l c c c c c c c@{}}
        \toprule
        & \multicolumn{2}{c}{{\bf vocab.}} & \multicolumn{3}{c}{{\bf augmentation}} & \multicolumn{2}{c}{\bf{score}} \\
        \cmidrule(lr){2-3}\cmidrule(l){4-6}\cmidrule(l){7-8}
        & random & corpus & color & ruby & bg & Acc. & NED \\ 
        \midrule
        Tesseract~\footref{footnote:tesseract} & \multicolumn{2}{c}{n/a} & \multicolumn{3}{c}{n/a} & 1.5 & 0.53 \\
        google cloud vision~\footref{footnote:gcv} & \multicolumn{2}{c}{n/a} & \multicolumn{3}{c}{n/a} & 21.5 & 0.35 \\ 
        \midrule
        \multirow{3}{*}{Ours w/o augmentation }
        &\cmark&      &      &      &      & 33.6 & 0.30 \\
        &      &\cmark&      &      &      & 38.4 & 0.28 \\
        &\cmark&\cmark&      &      &      & 39.3 & 0.28 \\
        \midrule
        \multirow{3}{*}{\textbf{Ours w/ augmentation}}
        &\cmark&\cmark&\cmark&      &      & 44.1 & 0.26 \\
        &\cmark&\cmark&\cmark&\cmark&      & 56.1 & 0.22 \\
        &\cmark&\cmark&\cmark&\cmark&\cmark& \textbf{61.7} & \textbf{0.19} \\
        \bottomrule
    \end{tabular}
\end{table*}
\begin{figure}[t]
    \includegraphics[width=1.0\linewidth]{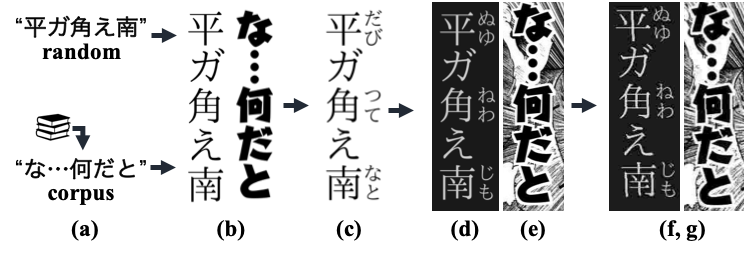}
    \caption{Procedures of rendering the training data.}
    \label{fig:synth}
\end{figure}

\subsection{Synthetic data generation}\label{sec:synth}
Noting the success of the synthetic dataset for scene text recognition~\cite{Jaderberg14c,Jaderberg16},
we decided to generate the training images synthetically.
We developed our own text rendering engine to generate text images optimized for manga.
Fig.~\ref{fig:synth} shows the rendering process and examples of synthetic text lines.
The steps below correspond to each process in Fig.~\ref{fig:synth} (a)--(g).
\setlength{\leftmargini}{15pt}
\begin{enumerate}
    \renewcommand{\labelenumi}{(\alph{enumi})}
    \item \textit{Text sampling.} The character sequence to be rendered is generated in two ways: sampling from the manga text corpus or randomly generating characters.
    By using a corpus built from manga, the model can implicitly learn the language model.
    However, this procedure cannot handle certain irregular patterns.
    Therefore, we decided to combine these two approaches, i.e., choosing 90\% from the corpus and 10\% at random.
    The text length is randomly chosen from 2 to 10.
    \item \textit{Font rendering.} A font is randomly selected from 586 and 1156 fonts for Japanese and English, respectively.
    The font size and weight are also varied randomly.
    \item \textit{Ruby.} Ruby, i.e., phonetic characters placed
    above Chinese characters, is added to 50\% of the data for the Japanese text.
    \item \textit{Coloring.} Foreground and background are filled with random colors.
    \item \textit{Background image composition.} Since the texts in manga are sometimes
    overlaid on the images, the images from the Manga109 dataset are used
    as the background images for 20\% of the data.
    \item \textit{Noise.} JPEG noise is added to the images.
    \item \textit{Distortion.} An affine transformation is used to distort the image.
\end{enumerate}
We generate five million of cropped line images with the processes above,
which are used to train the model described below.
As shown below,
each of the above processes helps to improve text recognition accuracy.

\subsection{Model}\label{sec:text_recog_model}
We follow the text recognition models introduced by Baek et al.~\cite{baek19}.
The images of the text line are resized to $50\times180$
and fed into the model.
The vertical text lines are rotated 90 degrees before resizing.
We tried the various combinations of modules described in \cite{baek19}
and found that the combination of
spatial transformer network~\cite{jaderberg2015},
ResNet backbone~\cite{cvpr_he2016},
Bi-LSTM~\cite{cheng2017focusing,shi2016robust,tpami_shi2017},
and attention-based sequence prediction~\cite{cheng2017focusing}
performed the best.

\subsection{Evaluation}
We evaluated the text recognition module with the annotated text in the Manga109 dataset.
Given each cropped speech bubble in the dataset,
we recognized the characters in the bubble using our text recognition module.
The metrics used here were the accuracy and normalized edit distance (NED)~\cite{karatzas2013icdar}.
The accuracy was computed as the number of correctly recognized texts (perfect match) divided by the total number of text regions (=12,542 regions).
Since there is no previous research on manga text recognition,
we compared with two existing OCR systems as baselines:
\textbf{Tesseract OCR},\footnote{\label{footnote:tesseract}\url{https://github.com/tesseract-ocr/tesseract}}, one of the most popular open-source OCR engines,
and
\textbf{Google cloud vision API},\footnote{\label{footnote:gcv}\url{https://cloud.google.com/vision/}}, a pre-trained OCR API provided by Google.

\begin{table*}
    \centering
    \caption{Text and frame detection performance on the Manga109 dataset}
    \label{tab:text_det}
        \begin{tabular}{@{}l l l l l l l@{}}
            \toprule
            & & & \multicolumn{2}{c}{{\bf text }} & \multicolumn{2}{c}{{\bf frame}} \\
            \cmidrule(l){4-5}\cmidrule(l){6-7}
            Method & \small{input image size} & backbone & $AP$ & $AP_{50}$ & $AP$ & $AP_{50}$ \\\midrule
            SSD-fork~\cite{corr_ogawa2018}
            & 300  & VGG            &  n/a & 84.1 & n/a  & 96.9 \\
            \midrule
            \multirow{6}{*}{Faster R-CNN}

            & 500  & ResNet-101     & 65.0 & 92.5 & 91.6 & 97.5 \\
            & 800  & ResNet-101     & 69.3 & 94.4 & 92.5 & 97.6 \\
            & 1170 & ResNet-101     & \textbf{71.2} & 94.9 & 92.5 & 97.7 \\
            & 1170 & ResNet-50      & 70.9 & 94.8 & 90.7 & 97.5 \\
            & 1170 & ResNet-101-FPN & 70.3 & 94.4 & 92.5 & 97.7 \\
            & 1170 & ResNeXt-101    & 70.4 & 94.5 & \textbf{92.9} & \textbf{98.5} \\
            \midrule
            RetinaNet         & 1170 & ResNet-101     & 70.6 & \textbf{95.4} & 89.8 & 98.3 \\
            \bottomrule
        \end{tabular}
\end{table*}

\begin{figure}[t]
    \includegraphics[width=1.0\linewidth]{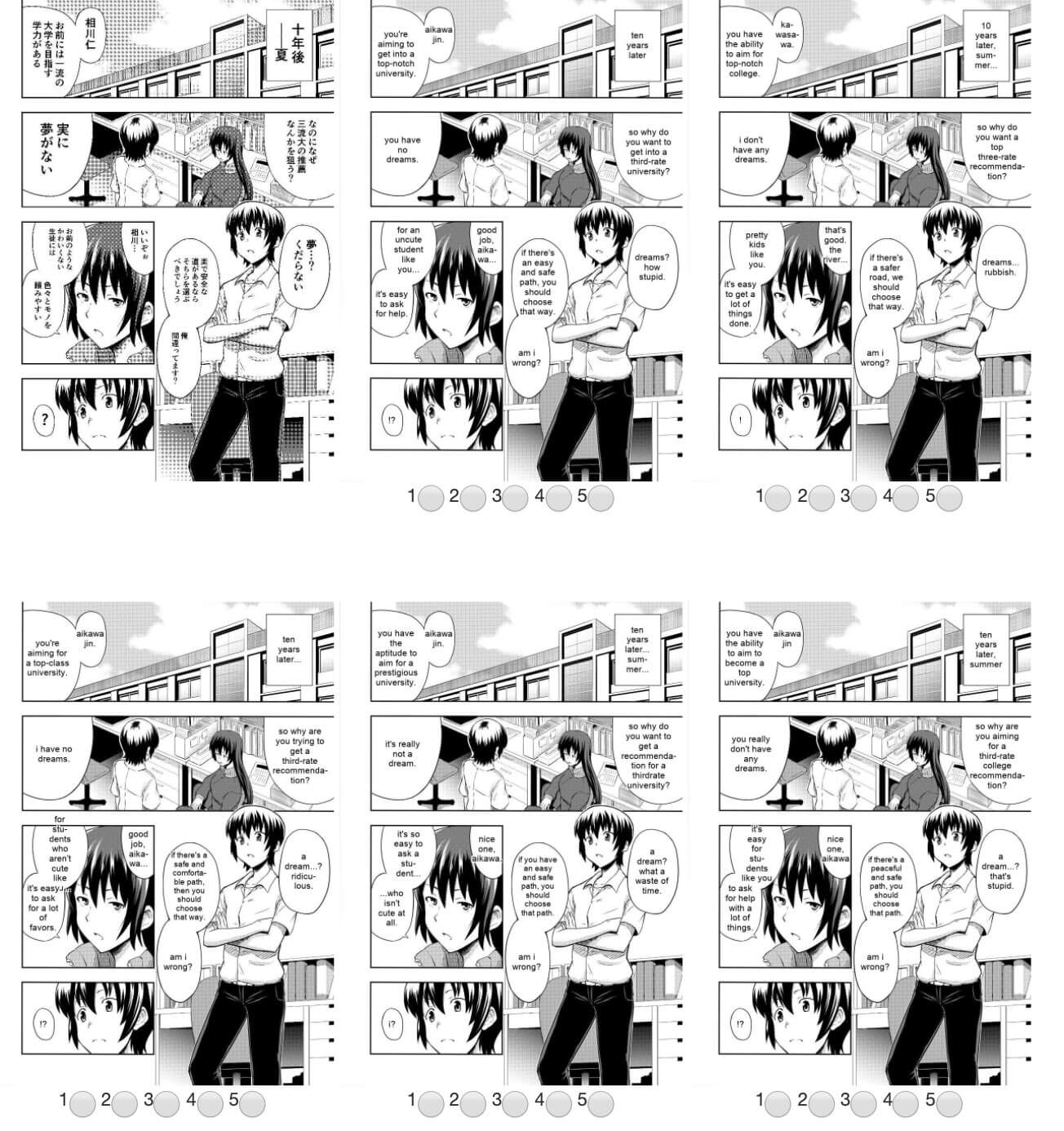}
    \caption{GUI of the user study. \copyright Mitsuki Kuchitaka}
    \label{fig:user_study_gui}
\end{figure}
\begin{figure*}[t]
    \begin{tabular}{cc}
        \begin{minipage}[t]{0.23\hsize}
            \vspace{1cm}
            \centering
            \includegraphics[width=1.0\linewidth]{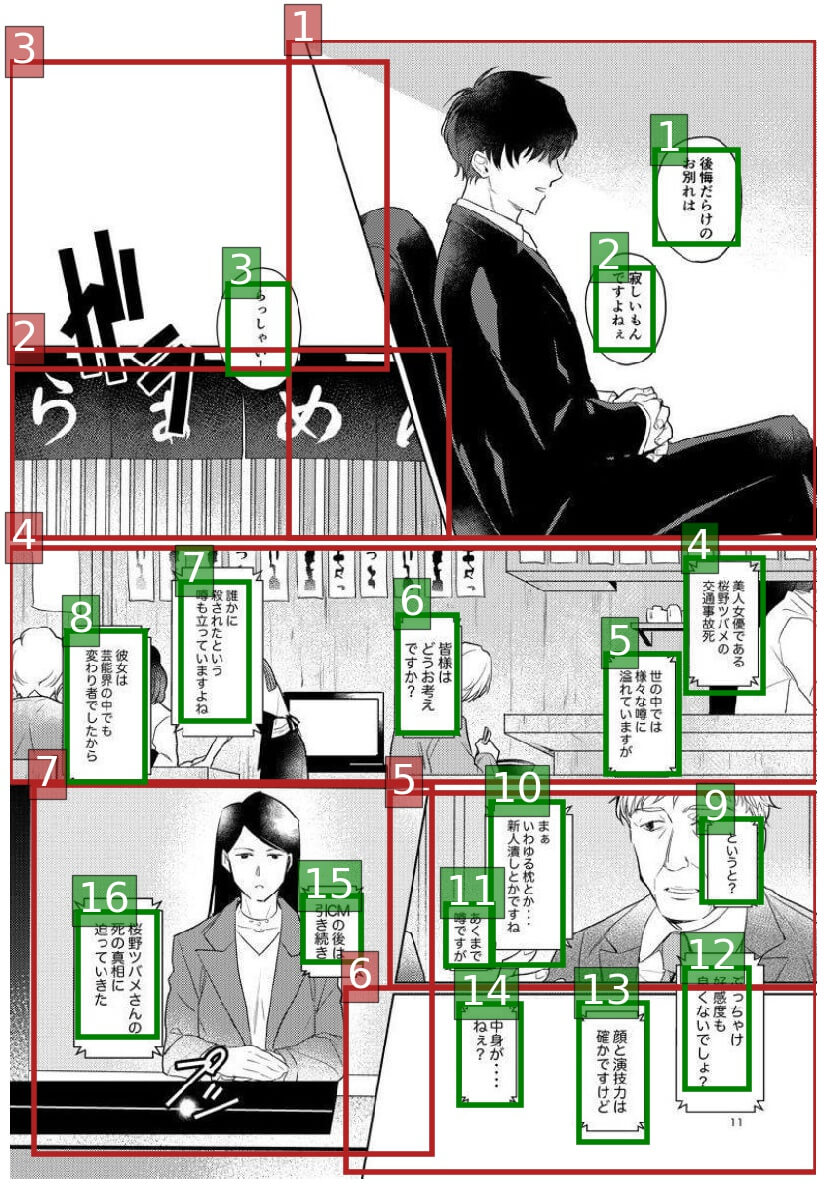}
            \subcaption{}
            \label{label1}
        \end{minipage}
        \hspace{0.2cm}
        \begin{minipage}[t]{0.23\hsize}
            \vspace{1cm}
            \centering
            \includegraphics[width=1.0\linewidth]{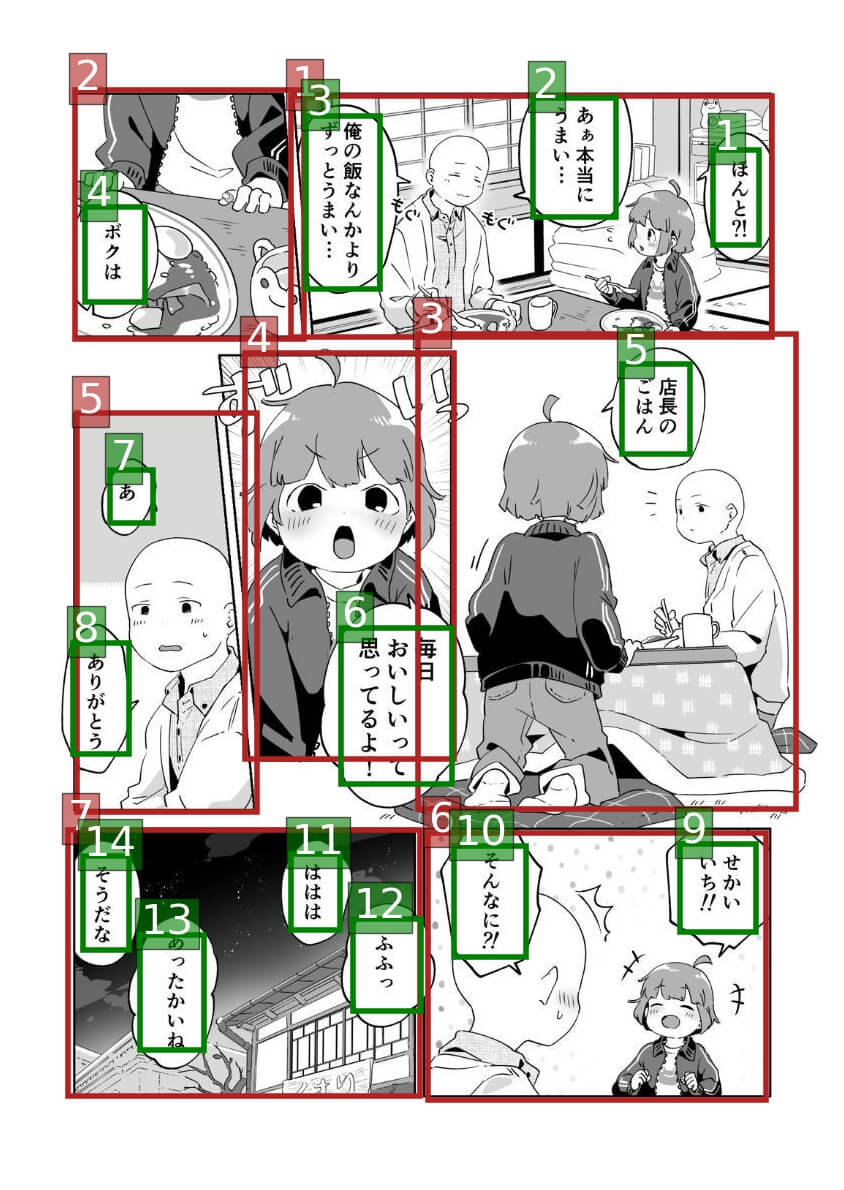}
            \subcaption{}
            \label{label2}
        \end{minipage}
        \hspace{0.2cm}
        \begin{minipage}[t]{0.23\hsize}
            \vspace{1cm}
            \centering
            \includegraphics[width=1.0\linewidth]{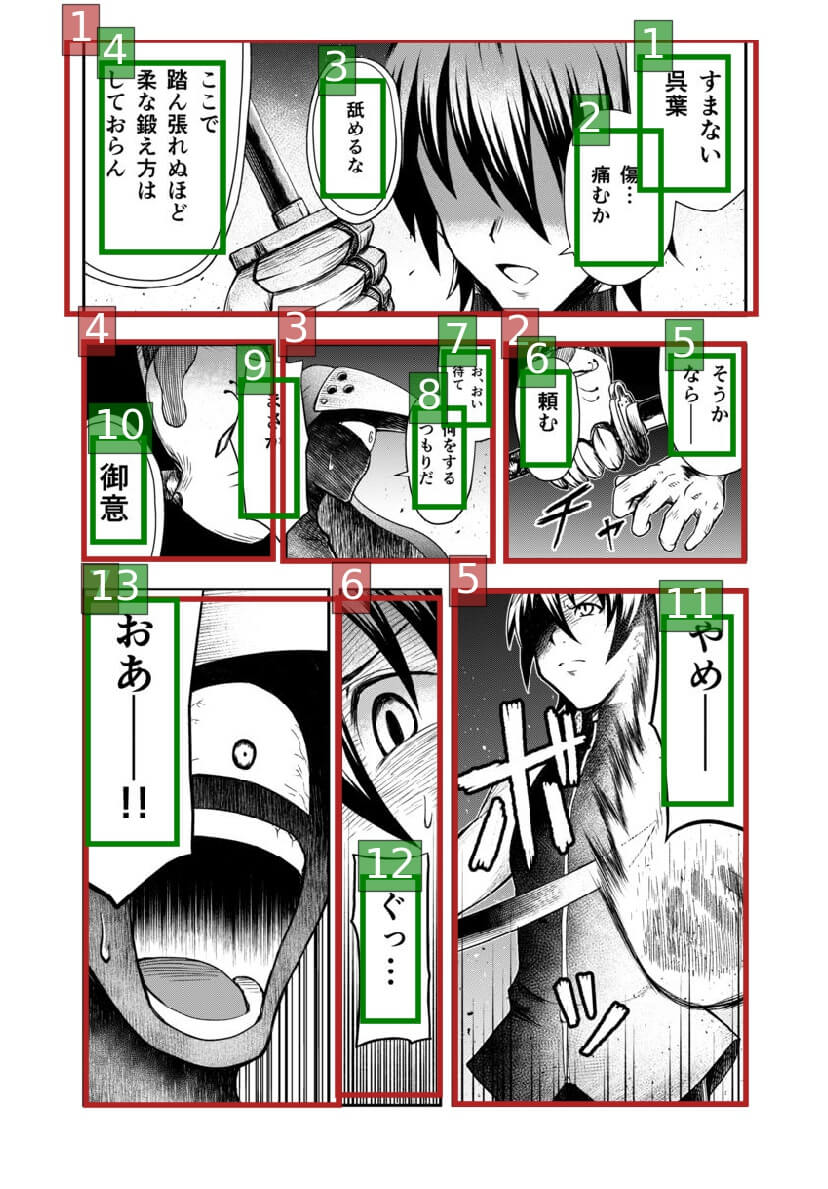}
            \subcaption{}
            \label{label1}
        \end{minipage}
        \hspace{0.2cm}
        \begin{minipage}[t]{0.23\hsize}
            \vspace{1cm}
            \centering
            \includegraphics[width=1.0\linewidth]{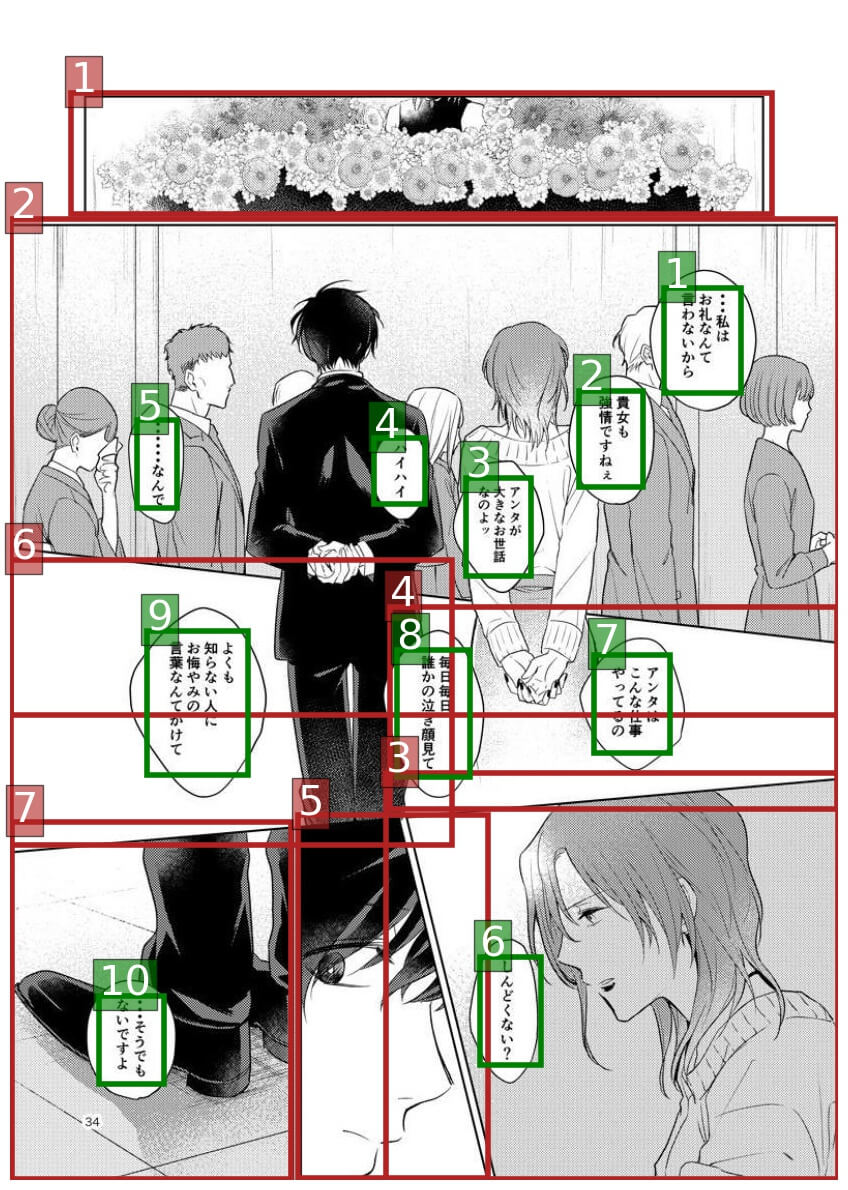}
            \subcaption{}
            \label{label2}
        \end{minipage}
    \end{tabular}
    \caption{Results of our text and frame order estimation.
    The bounding boxes of text and frame are shown in green and red rectangle, respectively.
    Estimated order of text and frame are depicted at the upper left corner of bounding boxes.
        \copyright Ito Kira \copyright Mitsuki Kuchitaka \copyright Nako Nameko}
    \label{fig:text_order}
\end{figure*}

Table~\ref{tab:text_recog} shows the text recognition performance.
Our method achieves much better accuracy than the existing systems.
This is because 1) we train the model using
images obtained by our manga text rendering engine,
and 2) the text line detection model optimized for manga can
properly discriminate the main characters and ruby characters.
We also performed an ablation study on our data augmentation process.
It demonstrated that every component significantly improved accuracy.
In particular, adding ruby characters and the background image
improved accuracy by 12.0 and 5.6 points, respectively.
Since ruby and the background image tended to be mistakenly recognized as part of a character,
our model learns to ignore them by adding them to the training data.

\section{Evaluation of Object Detection}
We here evaluate the performance of object detection.
We use a Faster R-CNN object detector to detect speech bubbles and frames in the manga image.
In accordance with Ogawa et al.~\cite{corr_ogawa2018},
we use Manga109 annotation to train and test our method,
where a speech bubble containing multiple text lines is considered as a single text area.
We use the average precision (AP) as an evaluation metric of this task.
$COCO AP$~\cite{corr_lin2014}, average AP for IoU from 0.5 to 0.95 with a step size of 0.05, and
$AP_{50}$ with a threshold of $\textrm{IoU}=0.5$ are computed.

Table~\ref{tab:text_det} shows the performance of object detection.
We compare our method with the one proposed by Ogawa et al.~\cite{corr_ogawa2018} because
it is the state-of-the-art method of text detection in Manga images.
Our system achieves significant improvements over the state-of-the-art method~\shortcite{corr_ogawa2018};
$84.1 \rightarrow \textbf{95.4}$ for text and $96.9 \rightarrow \textbf{98.3}$ for frame.
Although Ogawa et al.~\cite{corr_ogawa2018}
reported that the performance of Faster R-CNN is much poorer than that of their SSD-based model,
this is because they trained the Faster R-CNN as a multiclass object detector.
They mentioned that it is usually difficult to train a multiclass detection model on comic images
in the same way as a generic detection
because some objects are overlapping significantly.
Instead, we trained a Faster R-CNN with a single class.
The table also shows several important tips for object detection in manga.
For example, using a larger input size is effective for text classes, while it is not effective for frame classes because frames tend to be larger.
Therefore, in practice, the computational time can be reduced by using a small-sized input for the frame class.
In addition, several architectures that have had success in object detection tasks
(RetinaNet and ResNeXt/FPN backbone~\cite{lin2017focal,xie2017aggregated,lin2017feature})
does not improve the accuracy of this task.

\section{Hyperparameters of the NMT module}
We implement a Transformer (big) model with the fairseq~\cite{ott2019fairseq} toolkit and set its default parameters in accordance with~\cite{vaswani17}.
The model is trained using an Adam~\cite{kingma15} optimizer.
The hyperparameters of the model and optimizer are detailed in Tab.~\ref{tab:nmt_settings}.

\begin{table}[t]
    \caption{Hyperparameters for training the NMT model.}
    \label{tab:nmt_settings}
    \centering
    \begin{tabular}{@{}l l@{}}
        \toprule
        \# Layers of encoder & 6 \\
        \# Layers of decoder & 6 \\
        \# Dimensions of encoder embeddings & 1024  \\
        \# Dimensions of decoder embeddings & 1024  \\
        \# Dimensions of FFN encoder embeddings & 4096  \\
        \# Dimensions of FFN decoder embeddings & 4096  \\
        \# Encoder attention heads & 16 \\
        \# Decoder attention heads & 16 \\
        \midrule
        $\beta_1$ of Adam & 0.9 \\
        $\beta_2$ of Adam & 0.98 \\
        Learning rate & 0.001 \\
        Learning rate for warm-up & 1e-07 \\
        Warm-up steps & 4000 \\
        Dropout probability & 0.3 \\
        \bottomrule
    \end{tabular}
\end{table}

\section{GUI of User Study}
The evaluation system for the user study was developed as a web application.
The whole GUI is visualized in Fig.~\ref{fig:user_study_gui}.
A Japanese page and its English translated page are shown to a participant.
He/she selects the score for each sentence in the check box.
Unlike the usual plain text translation, this study directly compares the translated pages.

\section{More Examples of Text Ordering}
Examples of our text and frame order estimation are shown in Fig.~\ref{fig:text_order}.
Fig.~\ref{fig:text_order} (a)--(c) are successful cases; our approach
can correctly estimate the order of texts and frames in a manga image with complex structure.
Fig.~\ref{fig:text_order} (d) shows a failure case;
the system cannot handle some irregular cases such as the frames diagonally separated.
Such irregular cases should be detected and processed separately,
which has remained as future work.

\section{Text Cleaning Examples}
\begin{figure*}[h]
    \begin{tabular}{cc}
        \begin{minipage}[t]{0.22\hsize}
            \centering
            \includegraphics[width=1.0\linewidth]{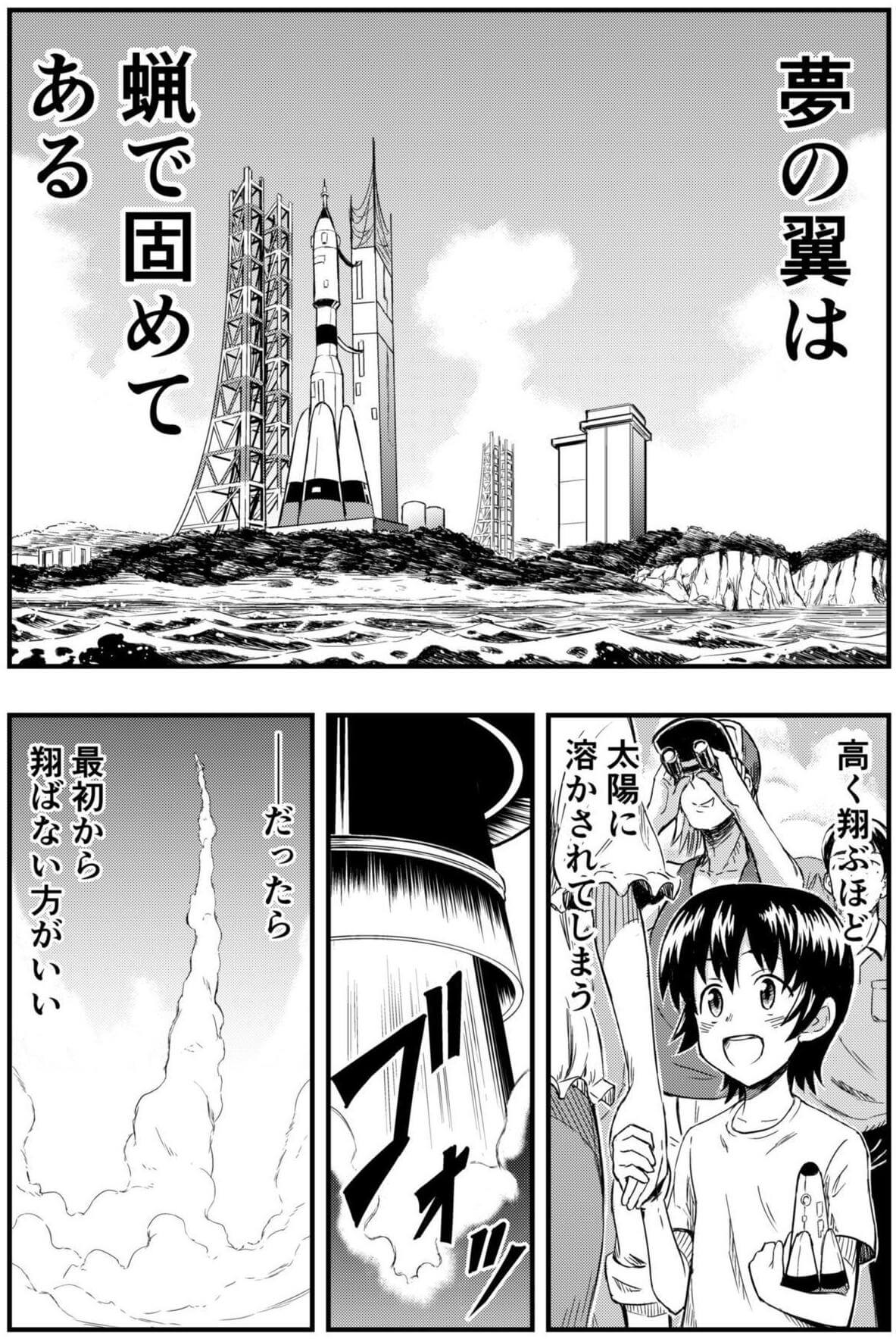}
            Original
            \label{label2}
        \end{minipage}
        \begin{minipage}[t]{0.22\hsize}
            \centering
            \includegraphics[width=1.0\linewidth]{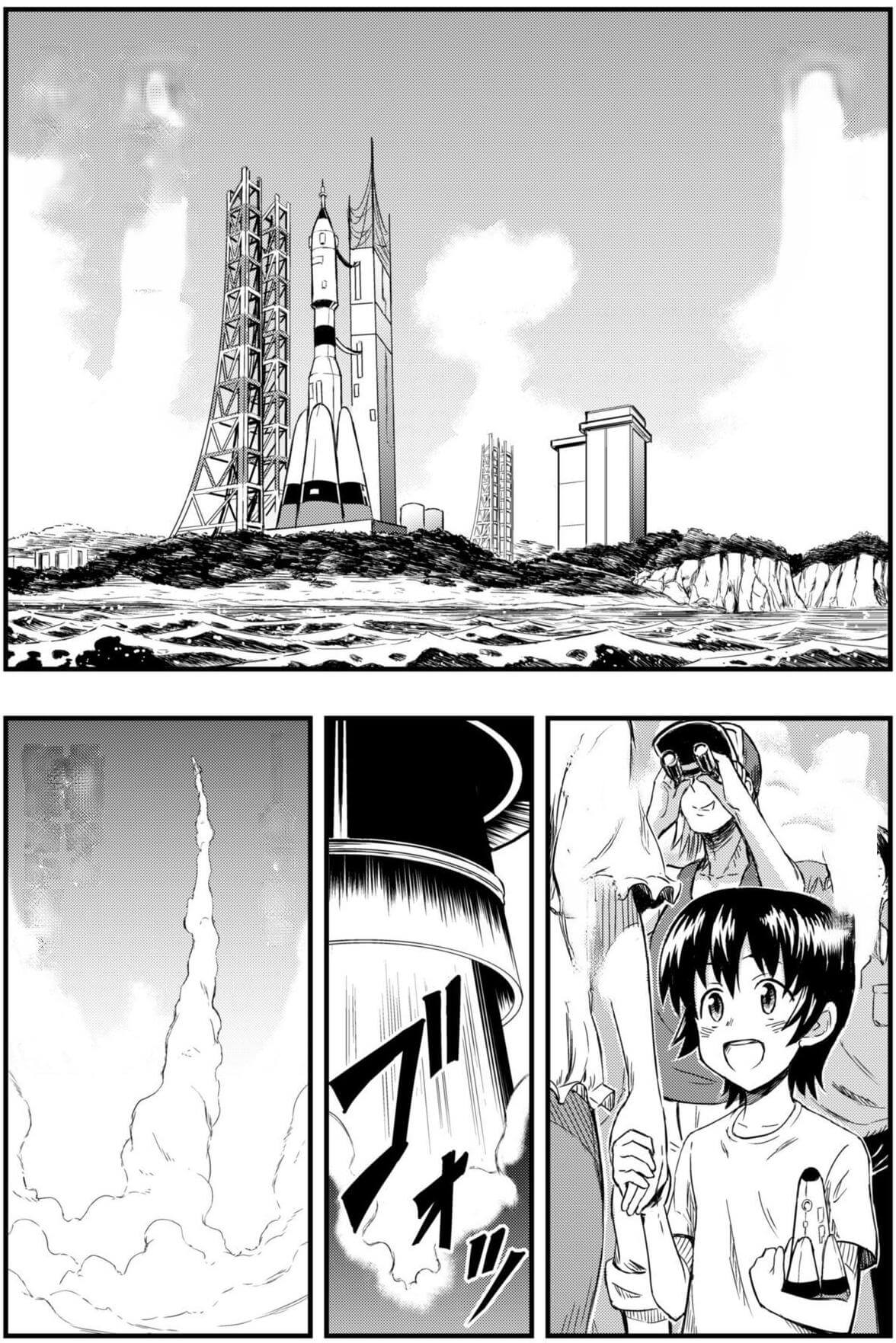}
            Cleaned
            \label{label2}
        \end{minipage}
        \hspace{1cm}
        \begin{minipage}[t]{0.22\hsize}
            \centering
            \includegraphics[width=1.0\linewidth]{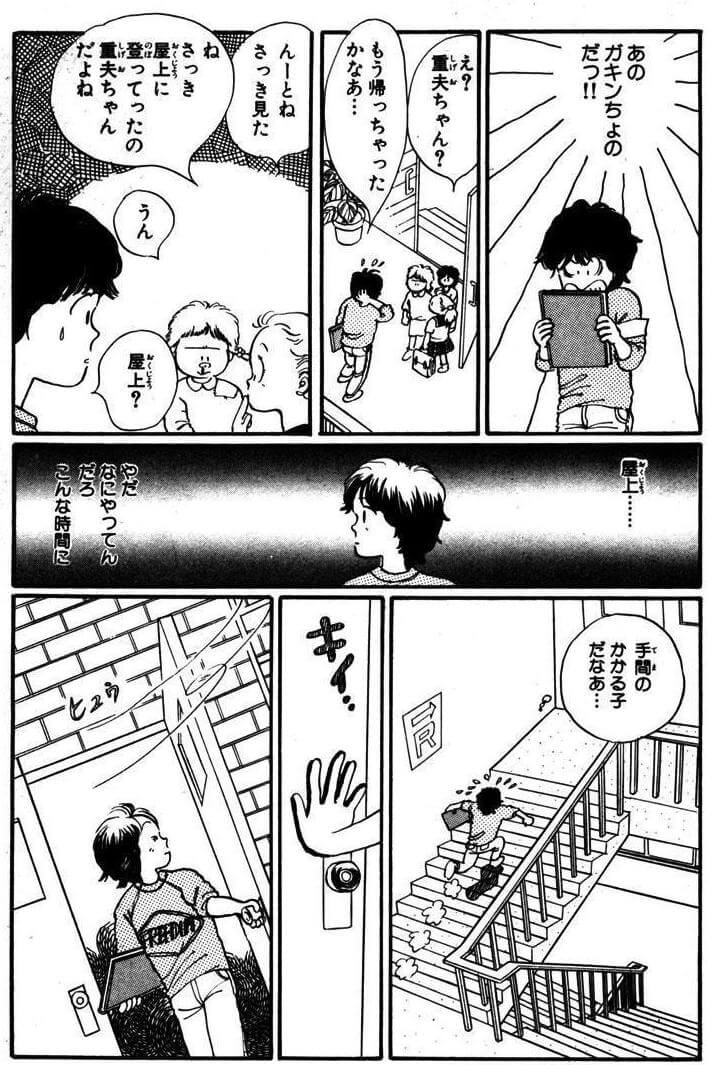}
            Original
            \label{label2}
        \end{minipage}
        \begin{minipage}[t]{0.22\hsize}
            \centering
            \includegraphics[width=1.0\linewidth]{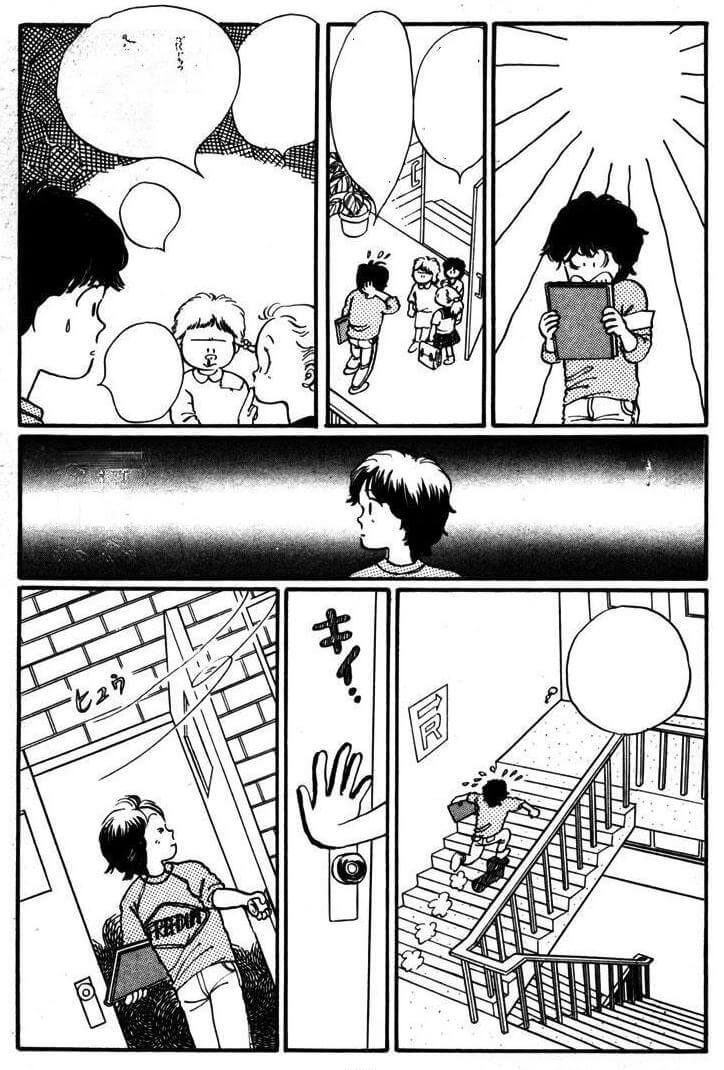}
            Cleaned
            \label{label2}
        \end{minipage}
    \end{tabular}
    \begin{tabular}{cc}
        \begin{minipage}[t]{0.22\hsize}
            \centering
            \includegraphics[width=1.0\linewidth]{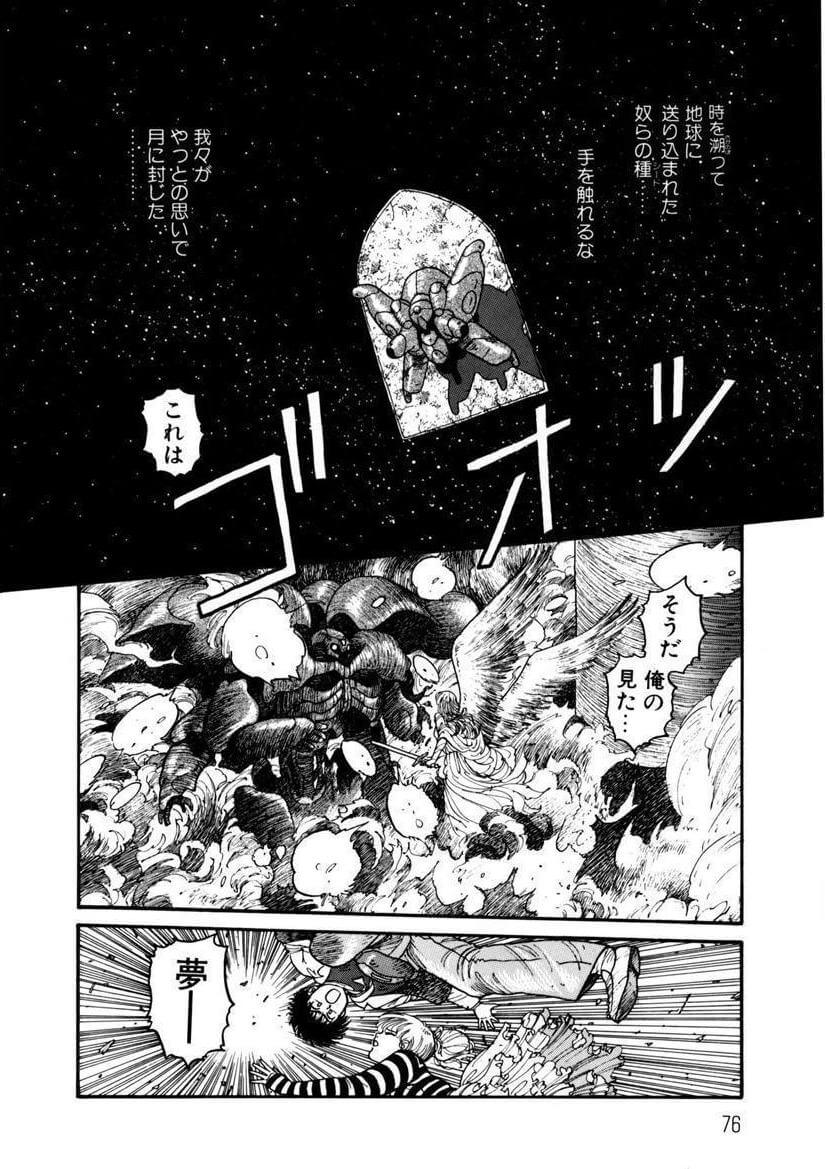}
            Original
            \label{label2}
        \end{minipage}
        \begin{minipage}[t]{0.22\hsize}
            \centering
            \includegraphics[width=1.0\linewidth]{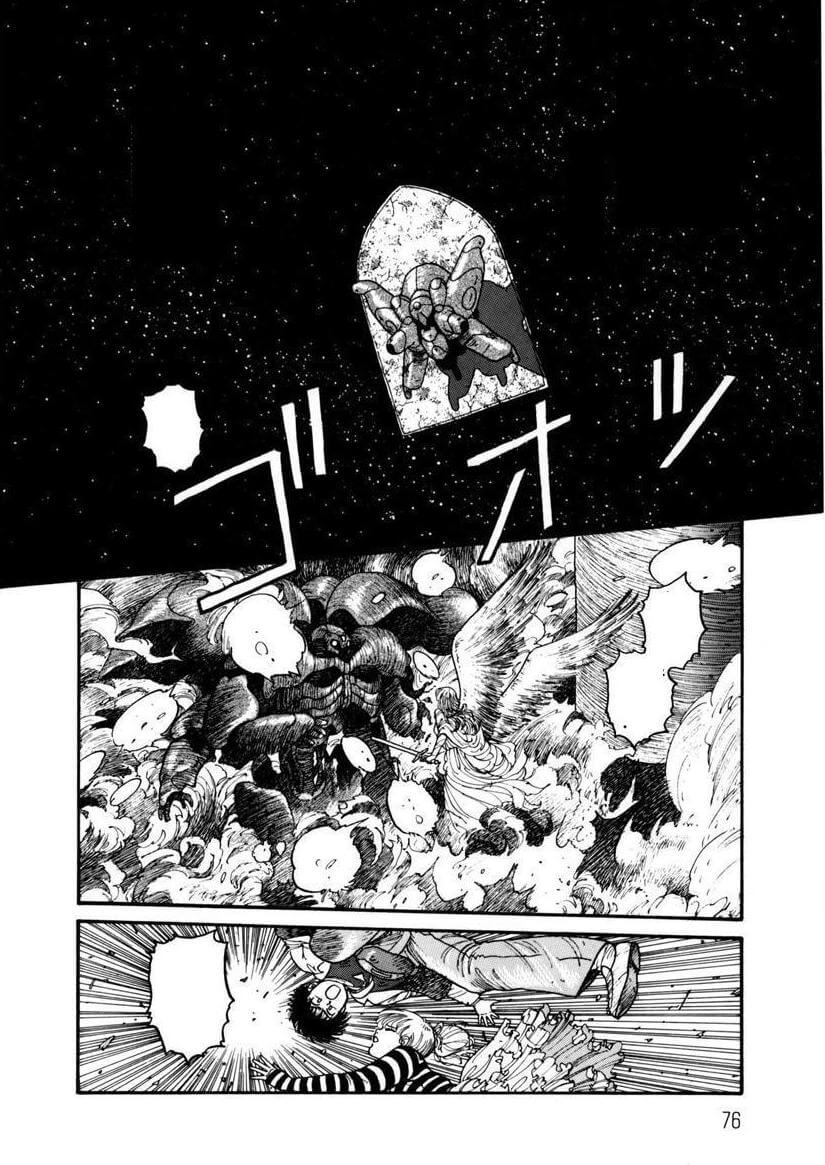}
            Cleaned
            \label{label2}
        \end{minipage}
        \hspace{1cm}
        \begin{minipage}[t]{0.22\hsize}
            \centering
            \includegraphics[width=1.0\linewidth]{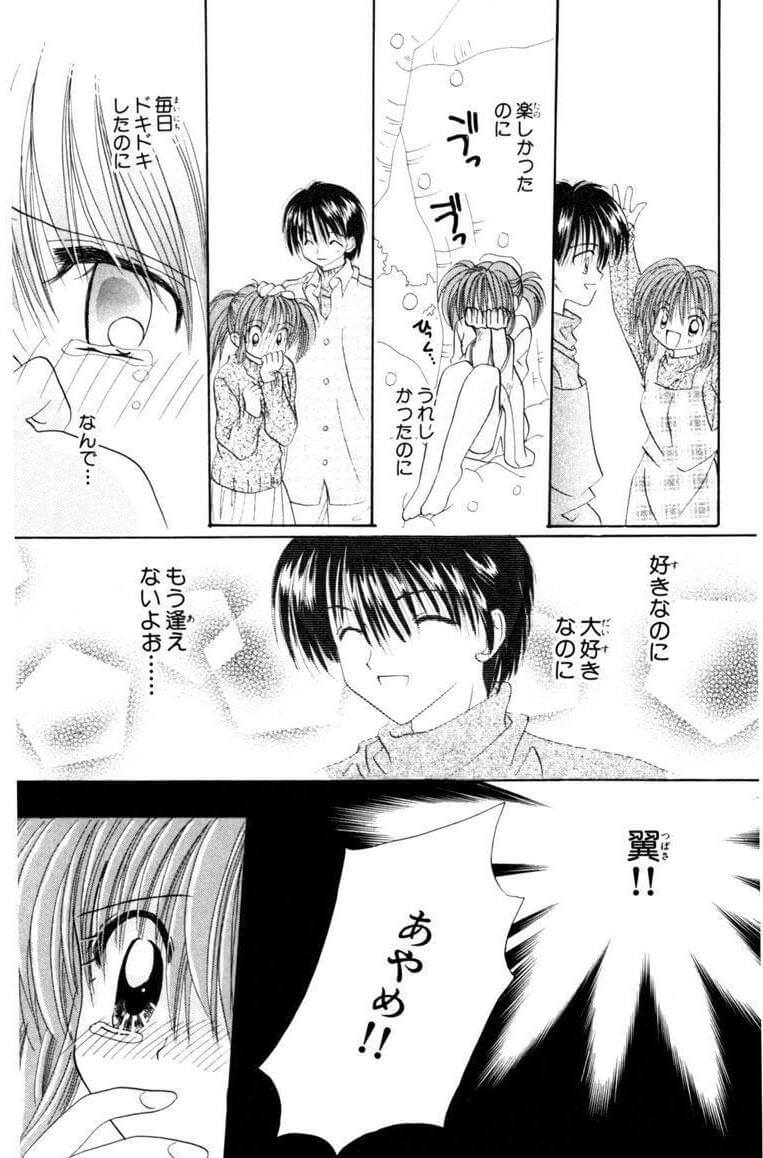}
            Original
            \label{label2}
        \end{minipage}
        \begin{minipage}[t]{0.22\hsize}
            \centering
            \includegraphics[width=1.0\linewidth]{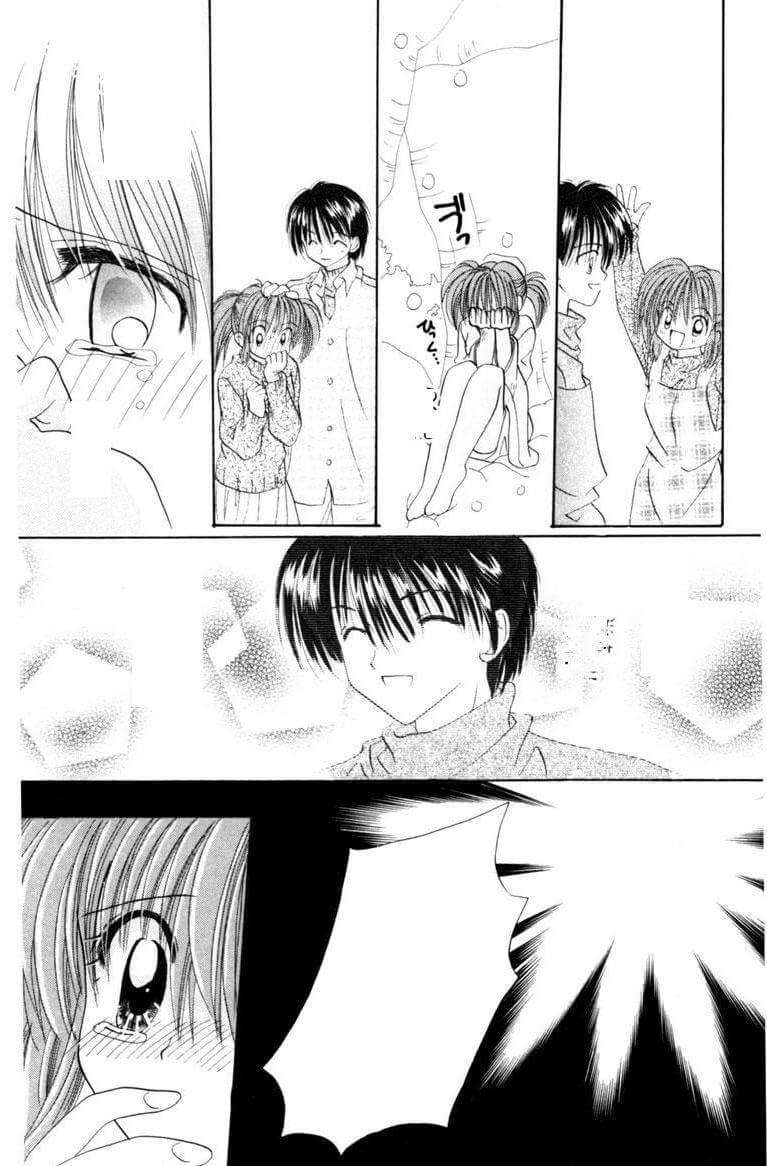}
            Cleaned
            \label{label2}
        \end{minipage}
    \end{tabular}
    \caption{Examples of text cleaning. \copyright Mitsuki Kuchitaka \copyright Masako Yoshi \copyright Masaki Kato \copyright Miki Ueda}
    \label{fig:text_clean_example}
\end{figure*}

Fig.~\ref{fig:text_clean_example} shows the examples of text cleaning.
Our inpainting-based method removes Japanese texts even if texts are on textures,
although the complemented texture is a little different from original one.

\section{More End-to-End Translation Examples}
\begin{figure*}[h]
    \begin{tabular}{cc}
        \begin{minipage}[t]{0.33\hsize}
            \vspace{1cm}
            \centering
            \includegraphics[keepaspectratio,scale=0.2]{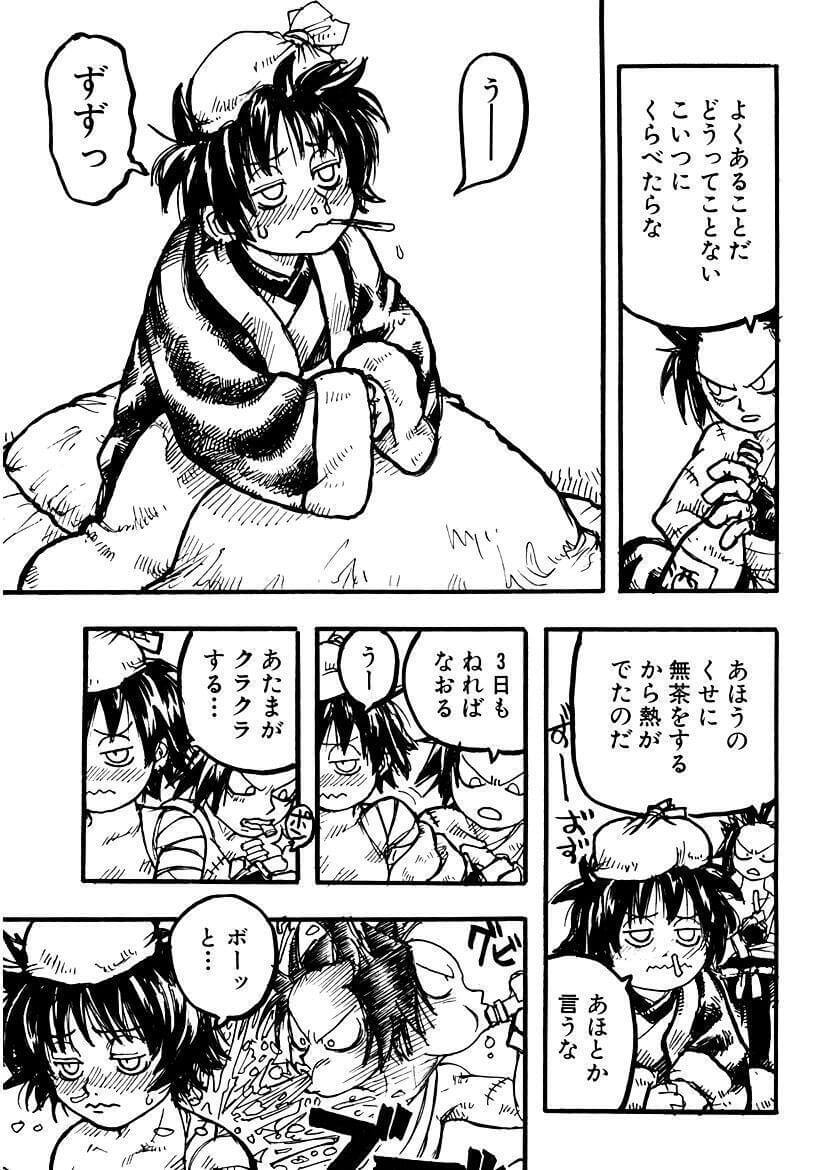}
            \subcaption{Original (Japanese)}
            \label{label1}
        \end{minipage}
        \begin{minipage}[t]{0.33\hsize}
            \vspace{1cm}
            \centering
            \includegraphics[keepaspectratio,scale=0.2]{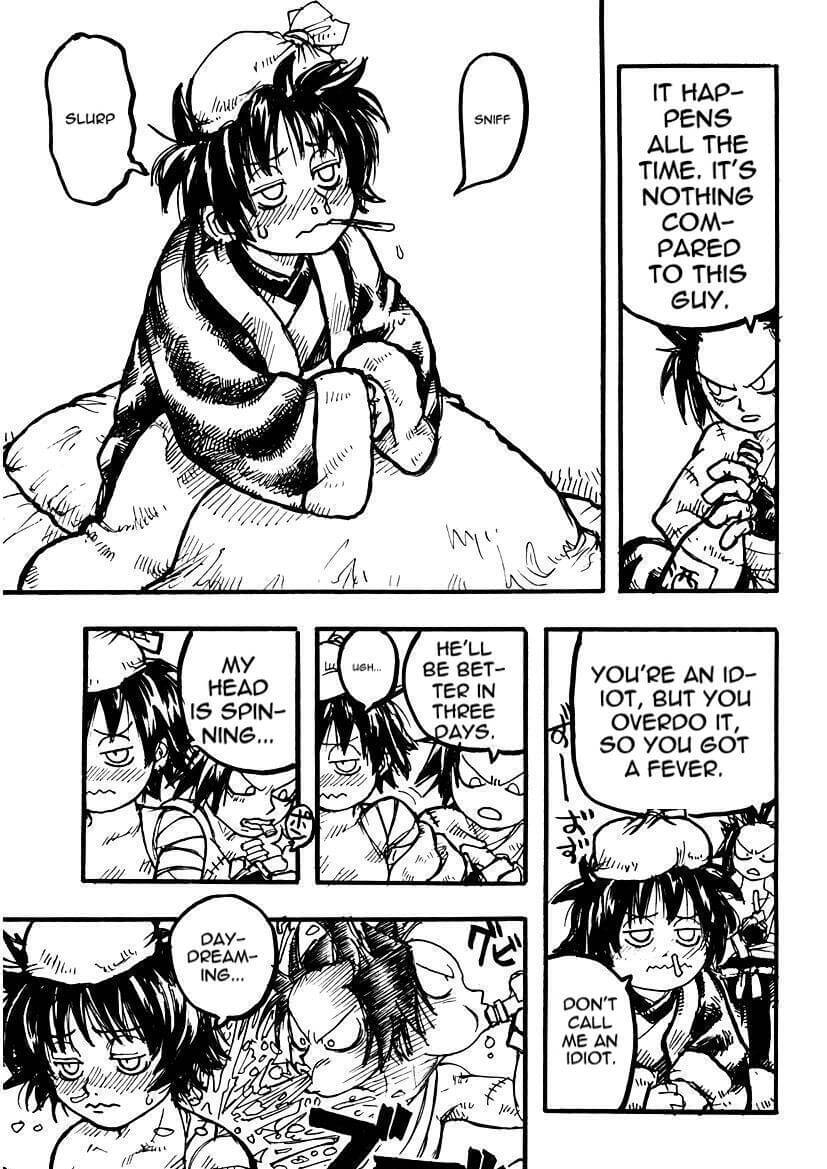}
            \subcaption{Translated (English)}
            \label{label2}
        \end{minipage}
        \begin{minipage}[t]{0.33\hsize}
            \vspace{1cm}
            \centering
            \includegraphics[keepaspectratio,scale=0.2]{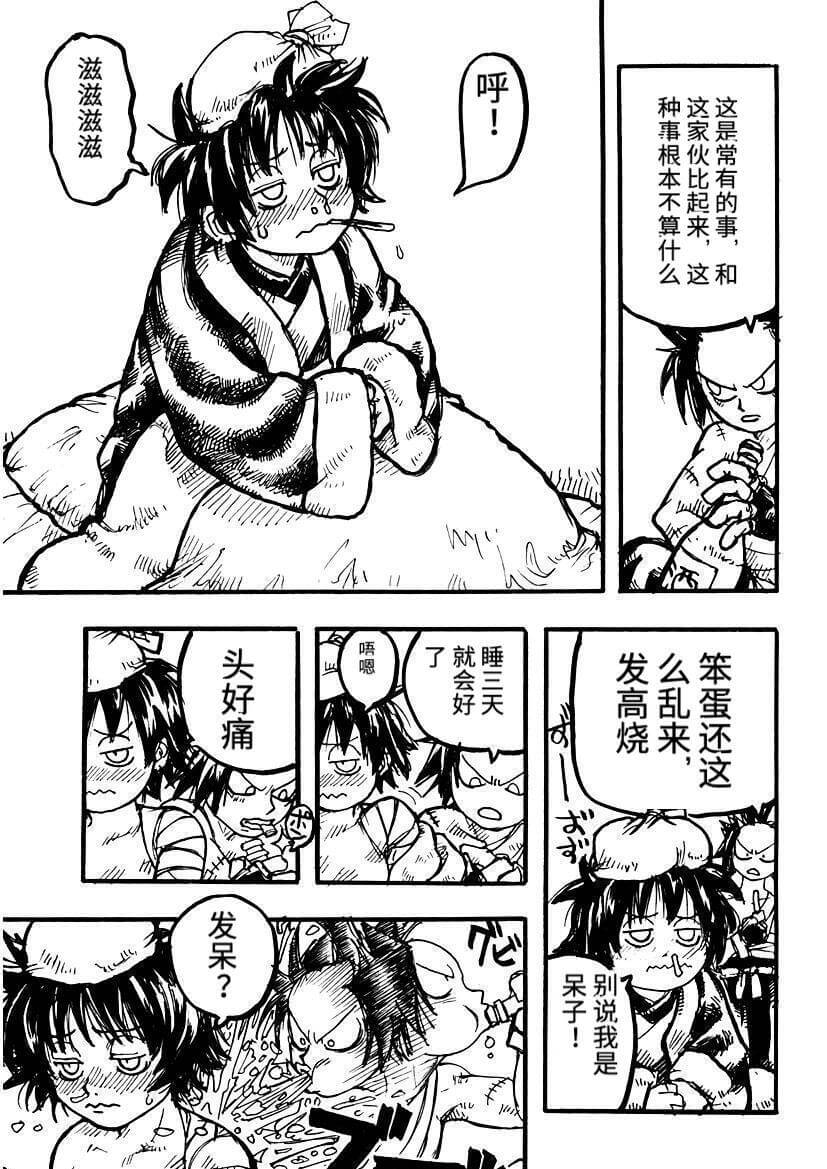}
            \subcaption{Translated (Chinese)}
            \label{label2}
        \end{minipage}
        \vspace{1cm}
    \end{tabular}
    \begin{tabular}{cc}
        \begin{minipage}[t]{0.33\hsize}
            \centering
            \includegraphics[keepaspectratio,scale=0.2]{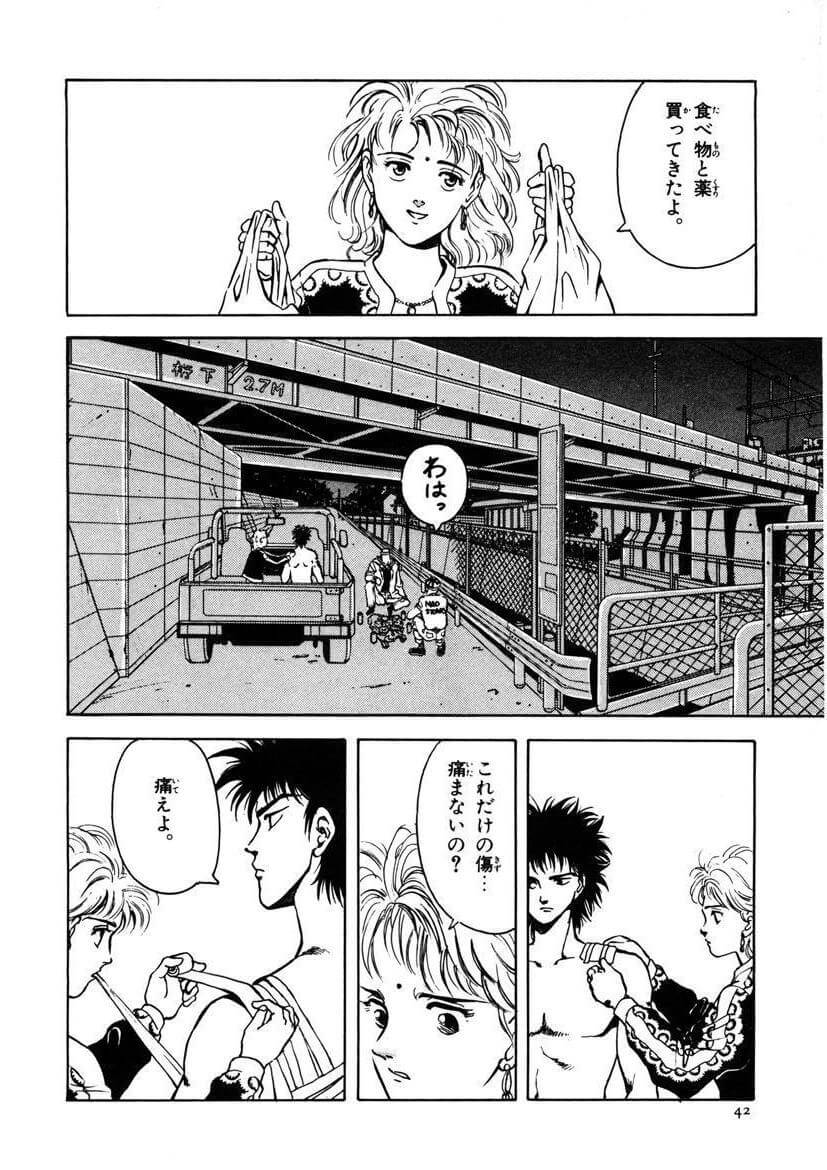}
            \subcaption{Original (Japanese)}
            \label{label1}
        \end{minipage}
        \begin{minipage}[t]{0.33\hsize}
            \centering
            \includegraphics[keepaspectratio,scale=0.2]{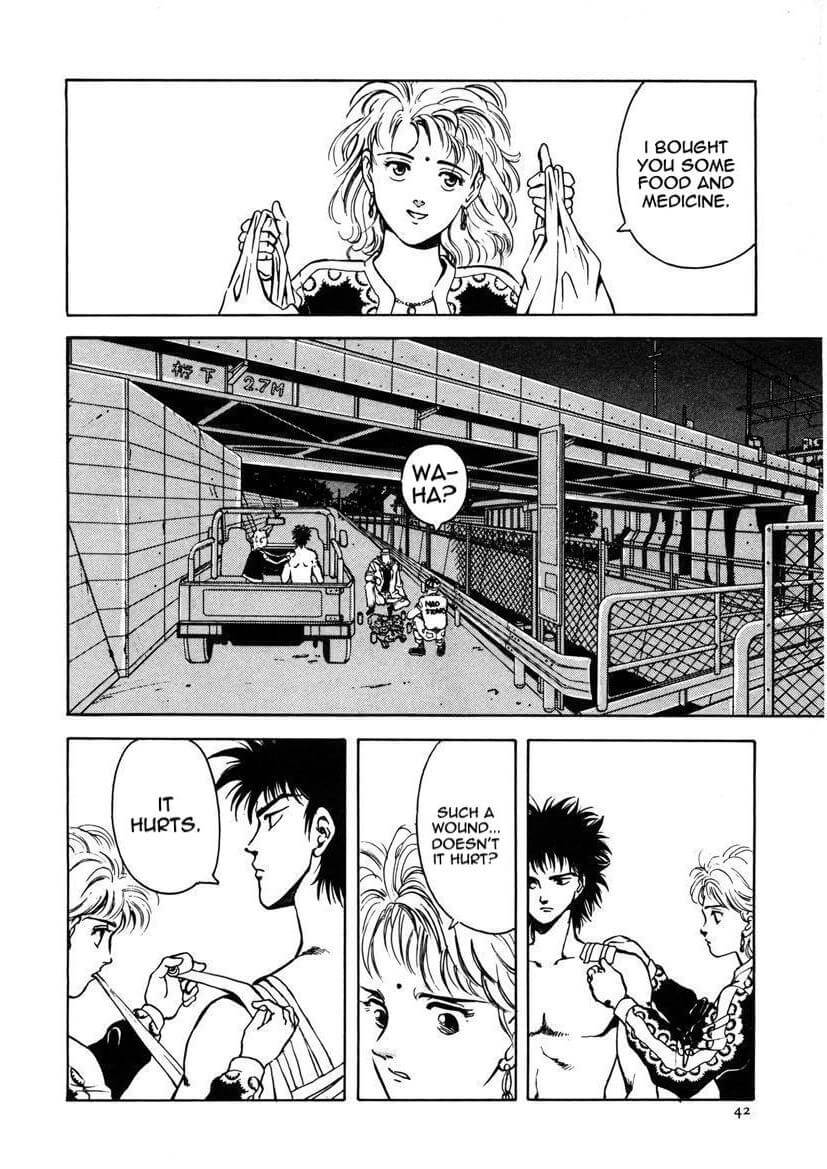}
            \subcaption{Translated (English)}
            \label{label2}
        \end{minipage}
        \begin{minipage}[t]{0.33\hsize}
            \centering
            \includegraphics[keepaspectratio,scale=0.2]{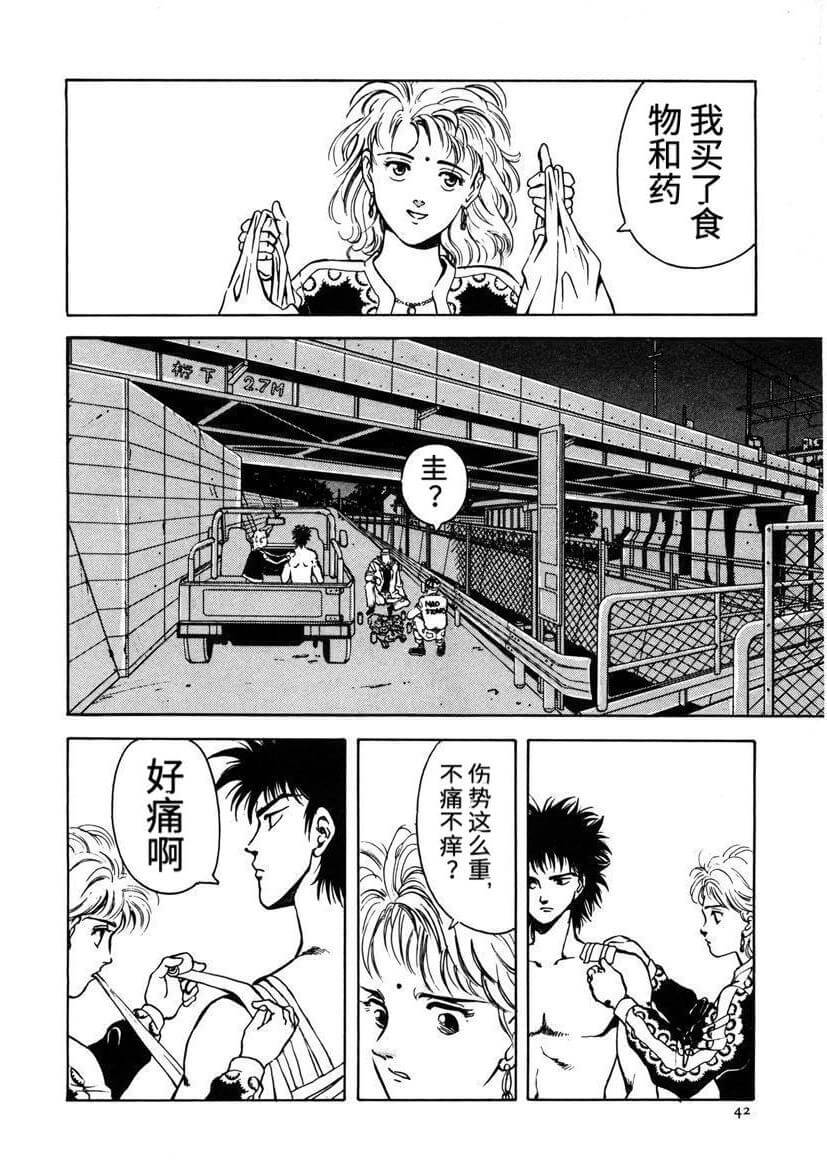}
            \subcaption{Translated (Chinese)}
            \label{label2}
        \end{minipage}
    \end{tabular}
    \caption{Examples of our translation. \copyright Syuji Takeya \copyright Hidehisa Masaki}
    \label{fig:e2e_example1}
\end{figure*}

\begin{figure*}[t]
    \begin{tabular}{cc}
        \begin{minipage}[t]{0.33\hsize}
            \vspace{1cm}
            \centering
            \includegraphics[keepaspectratio,scale=0.21]{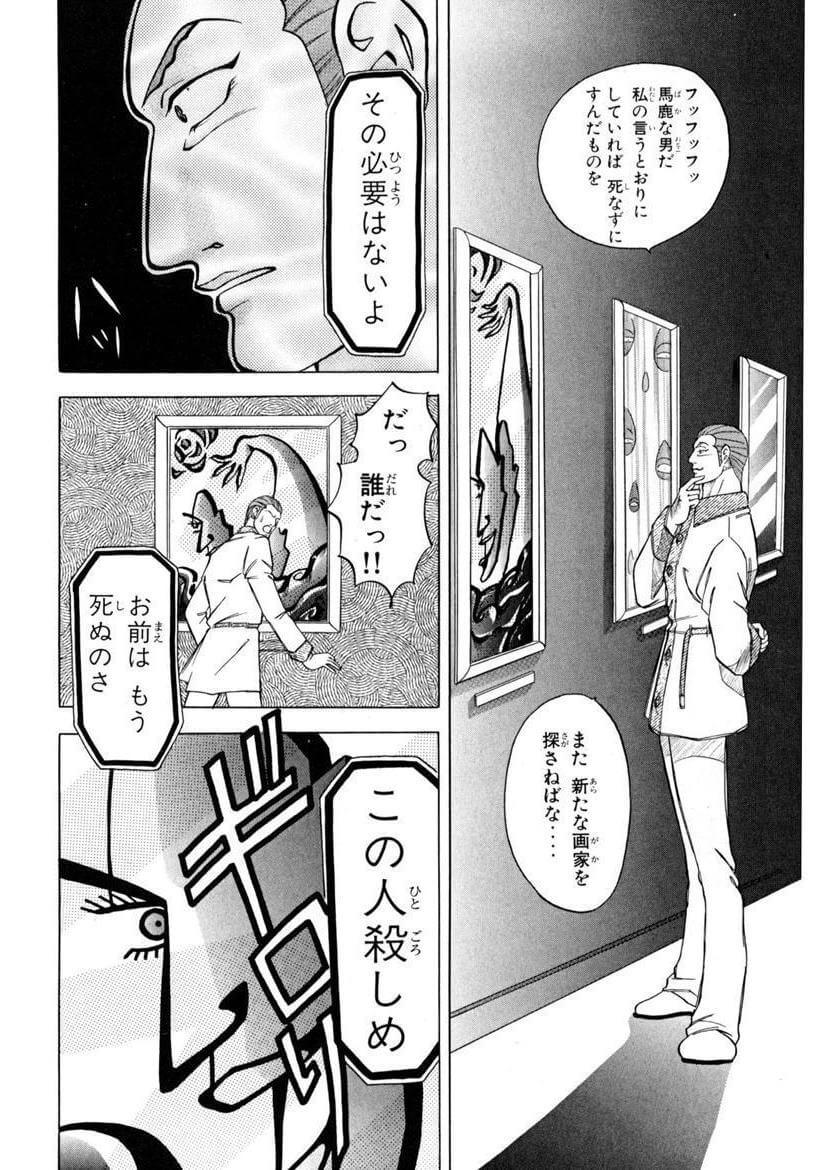}
            \subcaption{Original (Japanese)}
            \label{label1}
        \end{minipage}
        \begin{minipage}[t]{0.33\hsize}
            \vspace{1cm}
            \centering
            \includegraphics[keepaspectratio,scale=0.21]{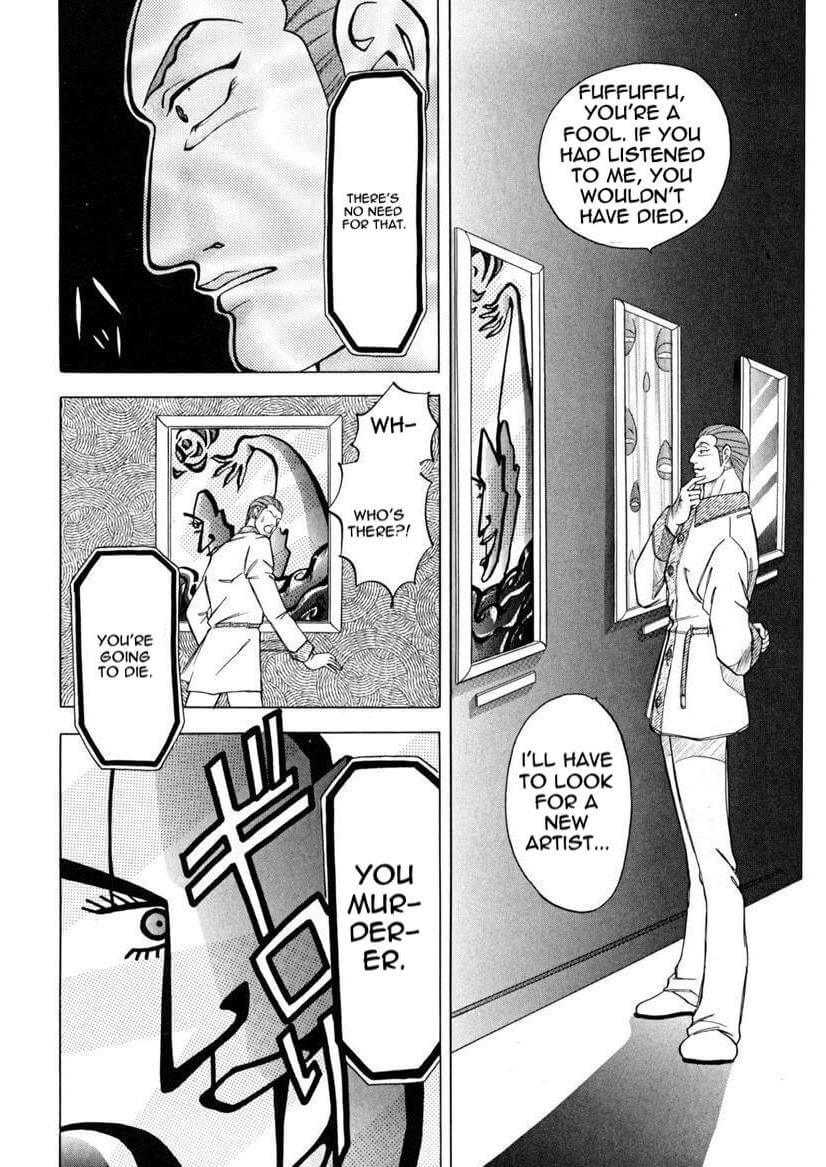}
            \subcaption{Translated (English)}
            \label{label2}
        \end{minipage}
        \begin{minipage}[t]{0.33\hsize}
            \vspace{1cm}
            \centering
            \includegraphics[keepaspectratio,scale=0.21]{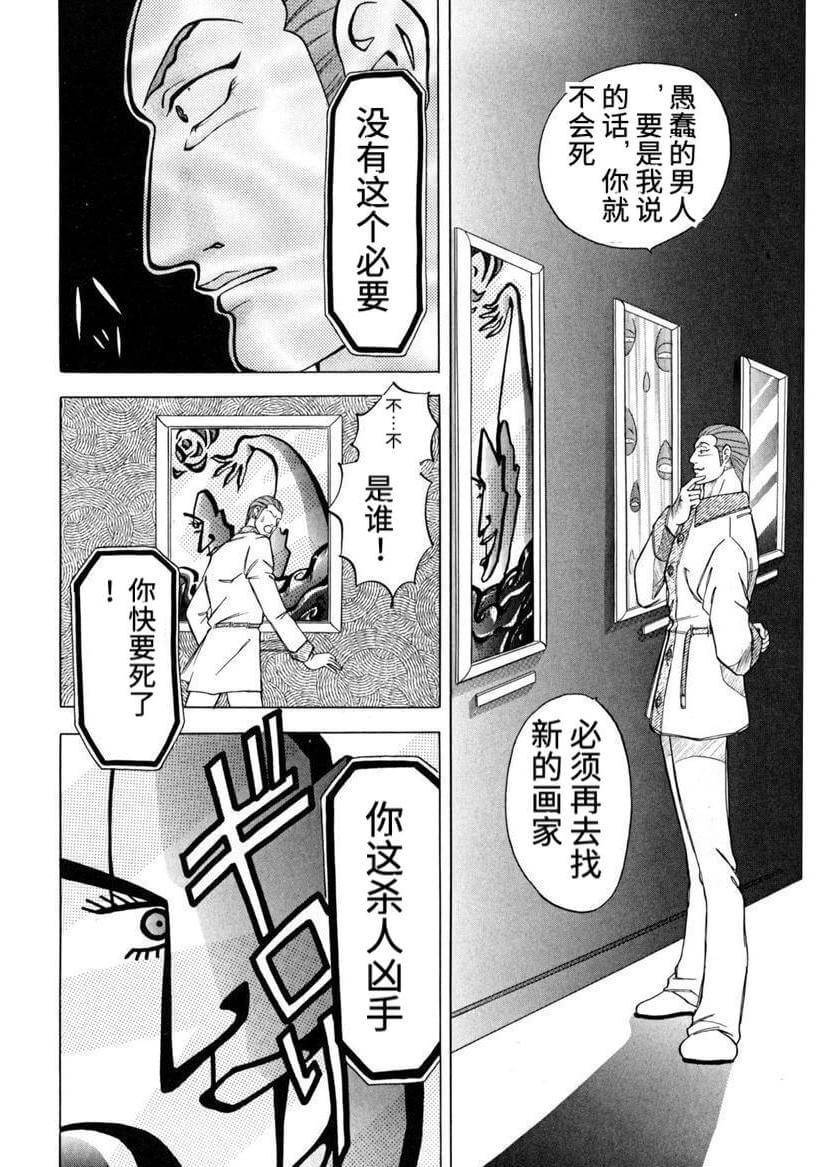}
            \subcaption{Translated (Chinese)}
            \label{label2}
        \end{minipage}
        \vspace{1cm}
    \end{tabular}
    \begin{tabular}{cc}
        \begin{minipage}[t]{0.33\hsize}
            \centering
            \includegraphics[keepaspectratio,scale=0.23]{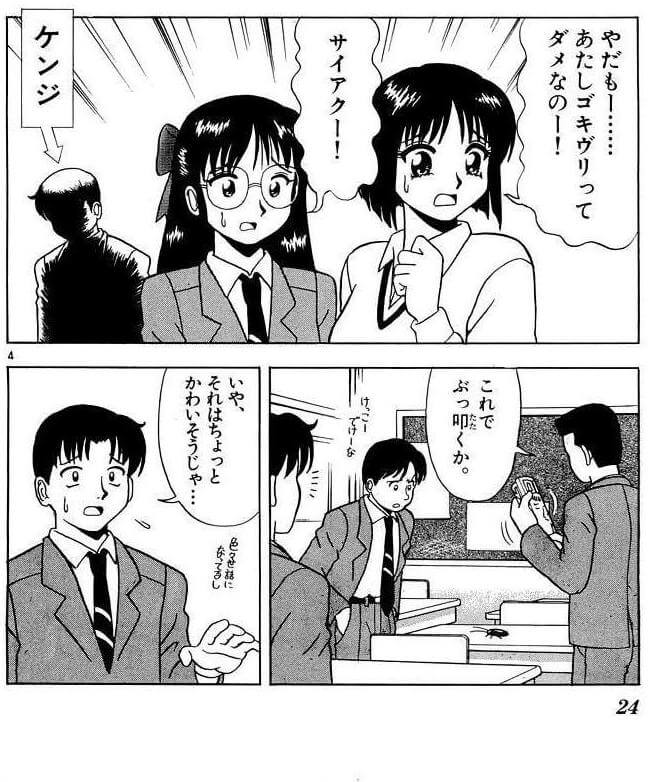}
            \subcaption{Original (Japanese)}
            \label{label1}
        \end{minipage}
        \begin{minipage}[t]{0.33\hsize}
            \centering
            \includegraphics[keepaspectratio,scale=0.23]{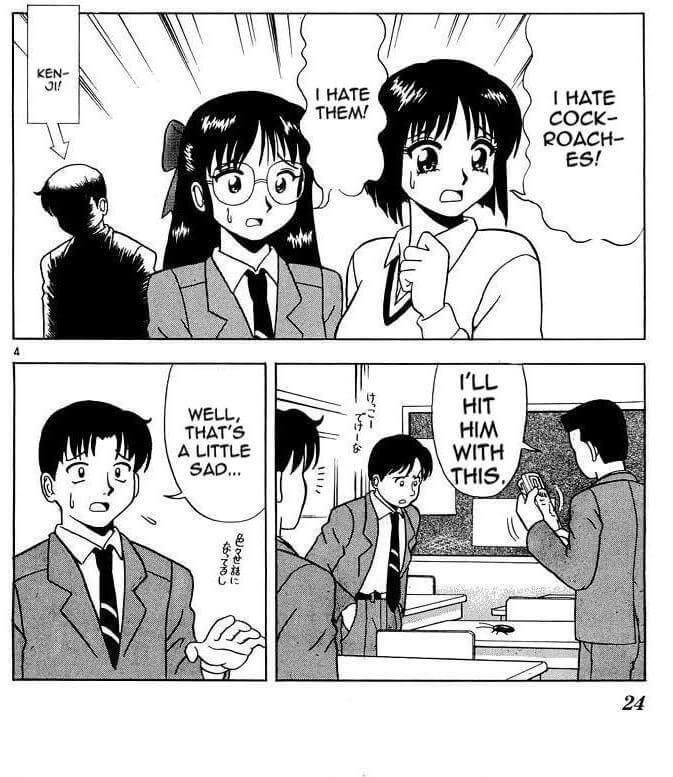}
            \subcaption{Translated (English)}
            \label{label2}
        \end{minipage}
        \begin{minipage}[t]{0.33\hsize}
            \centering
            \includegraphics[keepaspectratio,scale=0.23]{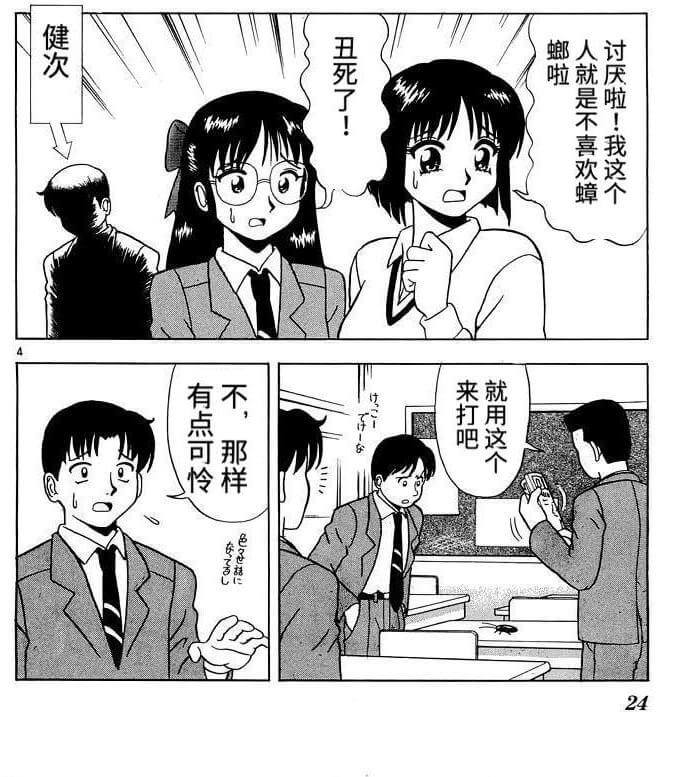}
            \subcaption{Translated (Chinese)}
            \label{label2}
        \end{minipage}
    \end{tabular}
    \caption{Examples of our translation. \copyright Masami Taira, \copyright Naoya Matsumori}
    \label{fig:e2e_example2}
\end{figure*}

Fig.~\ref{fig:e2e_example1} and \ref{fig:e2e_example2} shows more results of our fully automatic manga translation system.
The left images show the input pages, while the center and right figures show the translated results to English and Chinese.

\end{document}